\documentclass[
]{ceurart}

\usepackage{listings}
\usepackage{todonotes}
\usepackage{booktabs}
\usepackage{comment}

\newcommand{\thumb}[1]{\includegraphics[width=1.1cm,height=1.1cm]{#1}}

\makeatletter
\newcommand*{\rom}[1]{\expandafter\@slowromancap\romannumeral #1@}
\makeatother

\def\footnoterule{\relax%
  \kern-5pt
  \hbox to \columnwidth{\hfill\vrule width 0.5\columnwidth height 0.4pt\hfill}
  \kern4.6pt}
\makeatother

\usepackage{color,soul}
\sethlcolor{yellow}

\begin{document}
%
\copyrightyear{2024}
\copyrightclause{Copyright for this paper by its authors.
  Use permitted under Creative Commons License Attribution 4.0
  International (CC BY 4.0).}

\conference{ }

\title{Generative Outpainting To Enhance the Memorability of Short-Form Videos
}

%
\author[1]{Alan Byju}

\address[1]{School of Computing, Dublin City University, Glasnevin, Dublin 9, Ireland.}
\address[2]{Insight Research Ireland Centre for Data Analytics, Dublin City University, Glasnevin, Dublin 9, Ireland.}

\author[1]{Aman Sudhindra Ladwa}
\author[2]{Lorin Sweeney}[orcid=0000-0002-3427-1250]
\author[2]{Alan F. Smeaton}[orcid=0000-0003-1028-8389,email=Alan.Smeaton@DCU.ie,]
\cormark[1]

\cortext[1]{Corresponding author.}



%
\maketitle              
\begin{abstract}
With the expanding use of the short-form video format in advertising, social media, entertainment, education and more, there is a  need for such media to both captivate and be remembered. Video memorability indicates to us how likely a video is to be remembered by a viewer who has no emotional or personal connection with its content. This paper presents the results of using generative outpainting to expand the screen size of a short-form video with a view to improving its memorability. Advances in machine learning and deep learning are compared and leveraged to understand how extending the borders of video screensizes can affect their memorability to  viewers. Using quantitative evaluation we determine the best-performing model for outpainting and the impact of outpainting based on image saliency on video memorability scores.
\end{abstract}


\section{Introduction}

Within the current landscape of abundant available content the concept of video memorability is understood as the ability of an individual to recognise a video during subsequent viewings~\cite{Cohendet_2019_ICCV}. It is the sense of remembrance of visual content with whom a viewer has no  emotional connection or personal affiliations~\cite{sweeney2022overview}~\cite{DBLP:journals/corr/abs-2012-15635}. It is akin to viewing and remembering an advertisement, or a social media posting by another person with no real connection to the viewer.  

The cognitive underpinnings of memorability position it as a metric of cognition that is emotionally neutral and inherently determined by the content itself ~\cite{BAINBRIDGE2017141}, thereby making it resistant to a viewer’s distinctive emotional landscape or specific tastes. This highlights the idea that certain videos have inherent  qualities that make them more memorable, regardless of who the viewer is or their cognitive environment. 

A review of existing literature shows a scarcity of research directly targeting the manipulation of video memorability, despite a significant amount of research on its counterpart, image memorability. This gap is partly due to the complex nature of videos which, unlike static images, contain dynamic spatial-temporal elements. 
In this paper we attempt to alter memorability in short format videos by outpainting  using two different models and we evaluate the performance of these models on a standard benchmark video collection allowing us to compare our results with the results of others.

\section{Related Work}

\subsection{Image and Video Features and Memorability}

The concept of visual memorability encompasses multiple dimensions and in the context of images and videos, refers to the likelihood of recognising content upon subsequent viewings~\cite{sweeney2022overview}. Studies have shown that variations in image memorability are consistent across different viewers ~\cite{10.31234/osf.io/kd29q} indicating that memorability is an intrinsic part of the images.

The features of images and videos that are linked to memorability include local descriptors, spatial layout, and saliency. Local descriptors capture essential information about patterns, textures, shapes, or gradients in small patches of an image, enabling tagging and description of those patches. The Local Binary Pattern (LBP) is widely used for texture image classification and is applied in biometrics, object recognition, image retrieval, and enhancement~\cite{10.1007/978-981-16-1092-9_41}. Analysing spatial layout involves techniques like region segmentation, object detection, and depth estimation, which identify key components and their spatial interrelations. Finally, saliency refers to how distinct an object or area of an image is compared to its surroundings~\cite{wang2011image} and it is heavily used in many image processing and computer vision applications.

Computing the memorability of videos is now feasible with several techniques available.  A study analysing the memorability of video clips from the CSI TV series used a fine-tuned Vision Transformer architecture to predict memorability scores for scenes based on  aspects including character relationships \cite{cummins2022analysing}. Another study in the MediaEval Predicting Video Memorability Task in 2022 highlighted the importance of semantic information in predicting video memorability, correlating precise textual representations with higher memorability scores~\cite{guinaudeau2023textual}. However, active manipulation of  videos to influence their memorabilities are relatively new, with existing research focusing on prediction rather than enhancement of memorability. 

Because videos are dynamic and consist of sequences of frames that provide motion and temporal changes~\cite{10.11591/ijai.v9.i1.pp40-45} our approach here is to reconstruct  videos from individual frames to which we apply generative outpainting. For this we use techniques like autoencoders and recurrent neural networks (RNNs) as described later.

\subsection{Video Memorability Data Collections and MediaEval}



For video memorability, datasets like Memento10k and VideoMem facilitate research into why certain visual elements make videos more, or less memorable. 
The LAMDBA dataset released very recently consists of 2205 video ads appearing on YouTube with an average 33s duration and with long term memorability annotations from more than 1700 participants
~\cite{s2024longtermadmemorabilityunderstanding}.
Memento10k is a  dataset with 10,000 short videos, each manually annotated with memorability scores which was created and described in~\cite{newman2020multimodal}. 
The Memento10k dataset contains 10,000 videos with human annotations of memorability at different viewing delays. This dataset enables the modeling of video memorability decay over time, combining visual and semantic information. 
In addition to the manual video memorability annotations, each video also has a textual description of its contents. A comprehensive description of the dataset including links to download it, can be found at \url{http://memento.csail.mit.edu/}

Memento10k has been used for video storytelling and predicting media memorability, showcasing its broad applicability \cite{sweeney2021predicting}. Its high number of human annotations per video ensures reliable memorability scores, making it suitable for exploring video memorability.


One of the most significant catalysts to the development of computation methods to predict video memorability has been the MediaEval Multimedia Evaluation benchmark's ``Predicting Video Memorability" task. This challenges participants to develop algorithms for predicting short-term video memorability  using the Memento10K and VideoMem datasets~\cite{sweeney2022overview}. The tasks include video-based prediction using the Memento10K dataset, a second task to develop an understanding of a system’s adaptability with varied group data on the VideoMem dataset~\cite{sweeney2022overview} and a third task based on EEG data from subjects as they viewed  videos. The MediaEval benchmark has led to significant improvements in automatic memorability prediction for short-form videos \cite{de2020overview}, with applications in analysing TV series memorability \cite{cummins2022analysing} and enhancing memorability through saliency-based frame cropping~\cite{mudgal2024using}.

\begin{figure}[ht]
\centering
\includegraphics[width=0.9\linewidth]{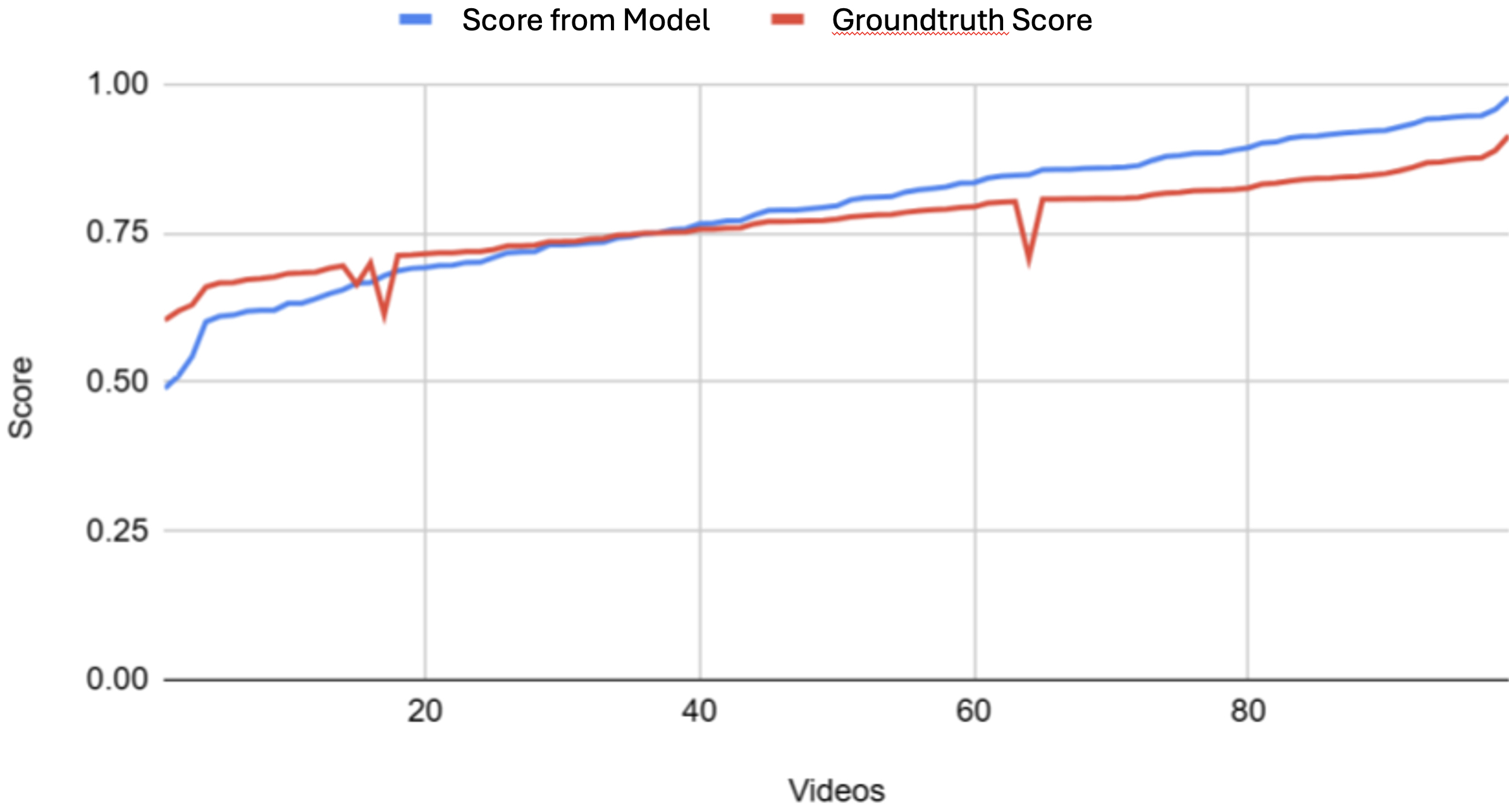}
\caption{Memorability scores from prediction model~\cite{sweeney2021predicting} vs. groundtruth scores for 100~X Memento10k  videos. Average score from  model = 0.788, average groundtruth score = 0.768
\label{fig:your_image_label}}
\end{figure}

As a basline for the work in this paper we use a submission to the benchmark  which uses a memorability score prediction model based on a Bayesian Ridge Regressor trained on CLIP features~\cite{sweeney2021predicting}. This model calculates a score between 0 and 1 for video frames, with higher scores indicating greater memorability, and computes the overall score for videos by aggregating frame scores. Figure~\ref{fig:your_image_label} shows the memorability scores using this model compared to the groundtruth  scores  provided with the dataset for a sample of 100 videos. Here we can see that the model scores memorability more conservatively than the groundtruth.

\subsection{Generative Image Models}

Among the prominent generative models used in multimedia exploration and augmentation are Diffusion Models, Generative Adversarial Networks (GANs), and Bayesian Networks. In this work  we focus more on Diffusion Models, which are foundational to the outpainting models used in our experiments.

The core advantage of diffusion models lies in their ability to create coherent and high-quality multimedia outputs. The process begins with a data sample to which noise is incrementally added until the sample is unrecognisable. The model then learns to reverse this process, step by step, removing the noise to reconstruct the original data or generate new, similar data. Although diffusion models require multiple forward passes during sampling, advances are being made to streamline this process \cite{fan2024hierarchical}. These models are particularly adept at adding depth and richness to videos, making them more engaging and memorable.

The application of diffusion models in video augmentation is transformative. Their ability to generate contextually rich backgrounds and elements can turn ordinary video content into captivating visual experiences. This capability is especially useful in enhancing video memorability as more compelling visuals encourage deeper viewer engagement~\cite{dhariwal2021diffusion}.  Recent developments have also shown that diffusion models can be guided to focus on specific features or styles, making them versatile for various multimedia applications. This adaptability is important for applications requiring high levels of detail and realism, such as inpainting and outpainting in video frames, where missing or damaged parts of the content need to be seamlessly recreated.


\subsection{Video Outpainting}

Inpainting and outpainting are image processing techniques that involve altering images in different ways. Image inpainting refers to the process of digitally altering an image in a manner that renders the adjustments imperceptible to a viewer lacking knowledge of the original image~\cite{10.35784/iapgos.5373} by filling in missing or damaged parts of an image with plausible information based on the surrounding content. Outpainting involves extending the content of an image beyond its original boundaries.

Video outpainting methods have challenges of maintaining consistency within and across frames. Flow-based methods such as optical flow, propagate pixel values into masked regions based on per-pixel motion between frames \cite{gao2020flowedge}~\cite{article}.  One of the earlier models that we explored was Dehan’s approach~\cite{article} even though it can produce poor foreground results due to its dependence on flow prediction, while being able to generate better backgrounds. 

More recent advances propose an approach for video outpainting using Latent Diffusion Models (LDMs) enhanced by a novel masked 3D diffusion model. This involves training a model to predict added noise to a video clip's latent representation while incorporating the video's contextual information. The goal is to reconstruct the original video from noise, handling each frame independently and later stitching them together. The masked 3D diffusion model (M3DDM)~\cite{fan2024hierarchical}  generates multiple frames simultaneously and maps video frames from pixel to latent space using a pre-trained encoder, replacing certain frames with raw video during training. This enables the model to use guide frames for inference, ensuring temporal consistency and improving generation quality through bidirectional learning. The training process includes a mask strategy that varies the masking of frame edges to support different outpainting scenarios. 

Finally, a very recent update in the field of video outpainting is  Mastering Video Outpainting Through Input-Specific Adaptation (MOTIA)~\cite{wang2024your}. This is a two-phase approach that combines input-specific adaptation with pattern-aware outpainting. This differs significantly from M3DDM as the input-specific adaptation phase allows MOTIA to learn unique patterns from a source video through a pseudo-outpainting task, optimising low-rank adapters inserted into a pre-trained diffusion model. The pattern-aware outpainting phase then leverages these learned patterns using spatial-aware insertion and noise regret techniques. MOTIA's architecture combines a variational autoencoder, CLIP text encoder, U-Net denoiser, temporal modules, ControlNet, and Blip for automatic captioning. 

MOTIA's difference to other approaches is particularly evident in out-of-domain scenarios, where M3DDM might struggle with unfamiliar video styles, while MOTIA can adapt on-the-fly \cite{wang2024your}. Additionally, M3DDM is limited to specific resolutions (256x256) and mask types (square), whereas MOTIA can handle arbitrary resolutions, video lengths, and mask types.

To compare the effectiveness of these three outpainting approaches, very recent work reported in~\cite{fan2024hierarchical} compares their performance on the Youtube VOS dataset~\cite{xu2018youtube}, a dataset consisting of of 4,453 videos spanning diverse scenes, categories and unique object instances. The authors applied the three  models to create horizontally outpainted videos and evaluated them based on visual differences and similarities between the original and reconstructed images using metrics including the Mean Squared Error, Peak Signal To Noise Ratio, Structural Similarity Index Measure, Learned Perceptual Image Patch Similarity and Frechet Video Distance.
Their results  showed M3DDM with significantly better performance over Dehan's method~\cite{article} and M3DDM also obtaining comparable results, but for different reasons~\cite{fan2024hierarchical}.  Since both M3DDM and MOTIA represent significant advances over previous video outpainting techniques, we shall include both in our own experiments in this paper.

\section{Experimental Methodology}
For our experiments we used the MOTIA and M3DDM models to generate outpainted videos on a subset of 100 videos randomly selected from the Memento 10K evaluation dataset. This contains a diverse collection of videos previously annotated with memorability scores. 
Table~\ref{tab:thumbnails} shows thumbnails of these 100 videos which included a variety of memorability scores  from low to high. Each selected video differed in scenes and unique object instances, allowing us to cover a large range of  video types from the original 1500 test videos.

\begin{table*}[htbp]
\centering
\caption{Thumbnail images of the 100 videos from Memento10k used in our evaluation.}
\begin{tabular}{|*{10}{c|}}
\hline
\renewcommand{\arraystretch}{1.5}

\thumb{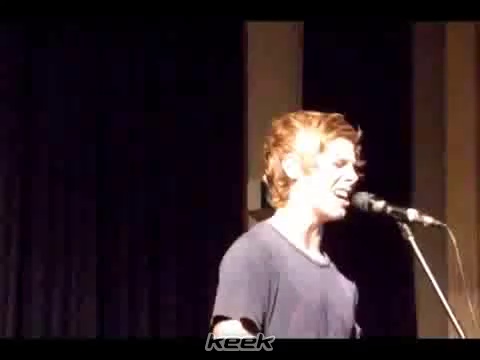} & \thumb{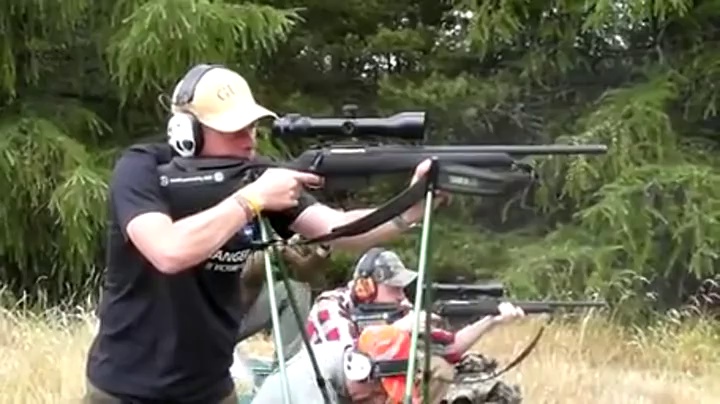} & \thumb{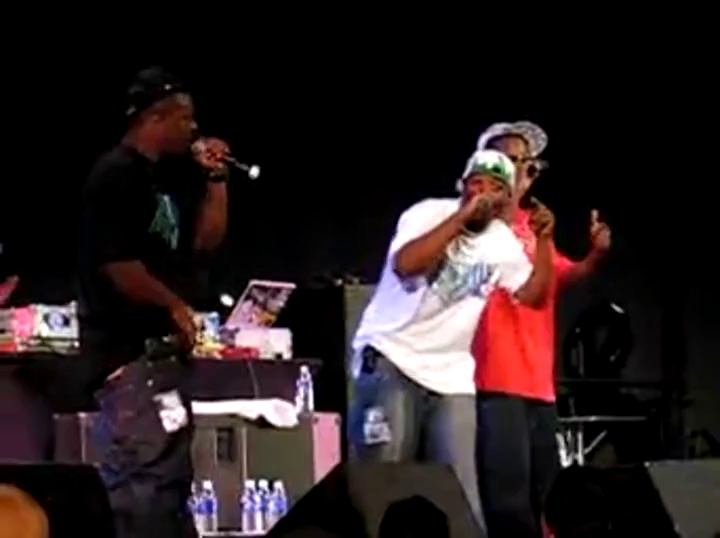} & \thumb{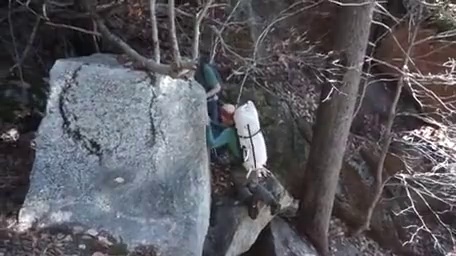} & \thumb{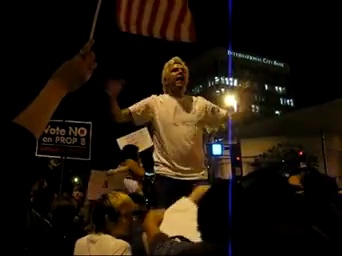} & \thumb{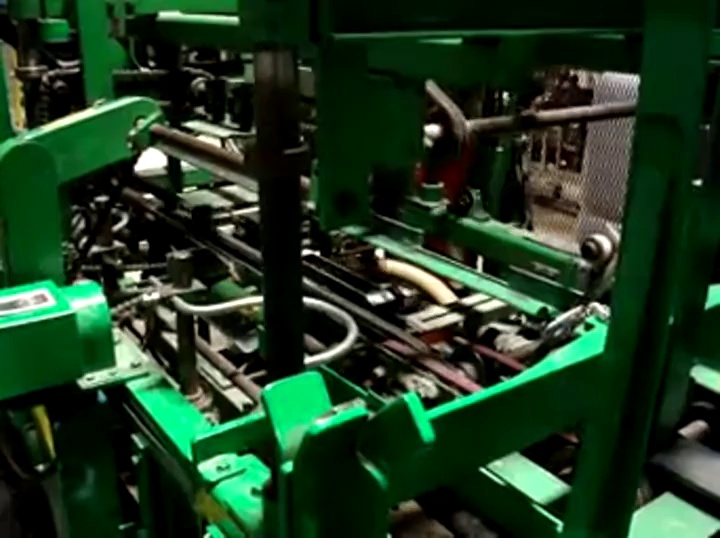} & \thumb{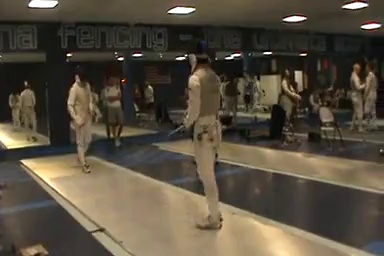} & \thumb{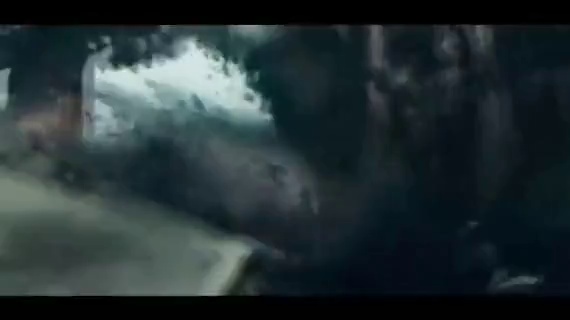} & \thumb{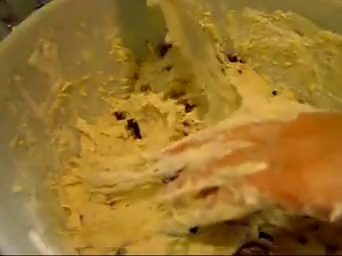} & \thumb{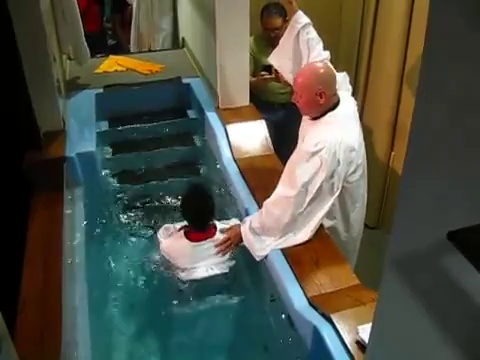} \\
\hline
\thumb{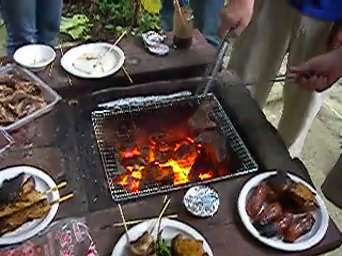} & \thumb{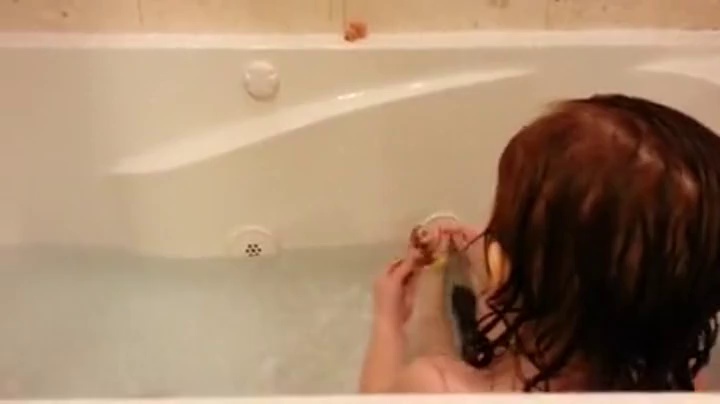} & \thumb{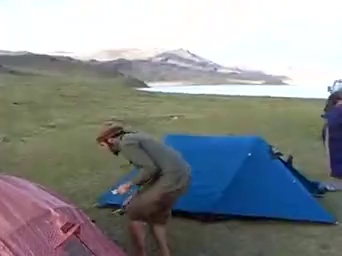} & \thumb{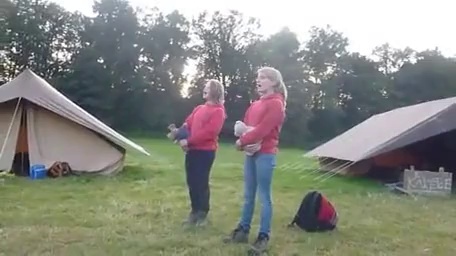} & \thumb{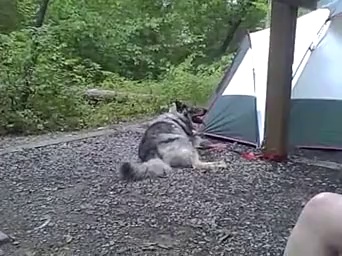} & \thumb{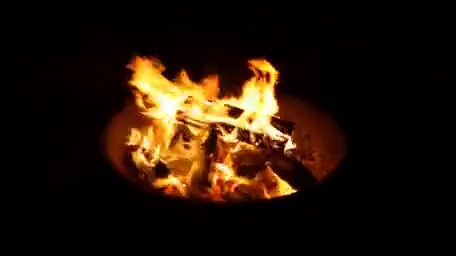} & \thumb{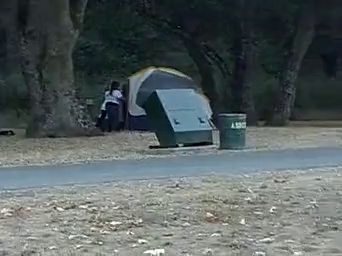} & \thumb{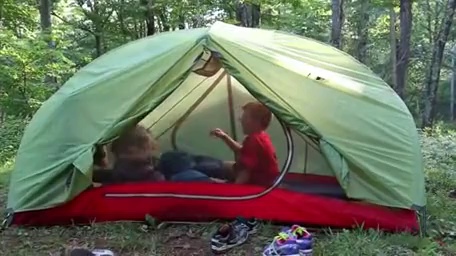} & \thumb{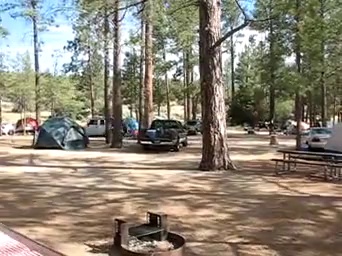} & \thumb{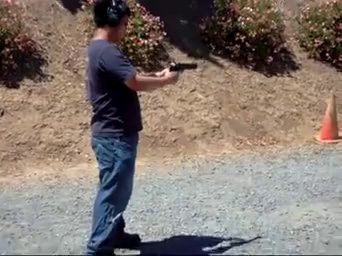} \\
\hline
\thumb{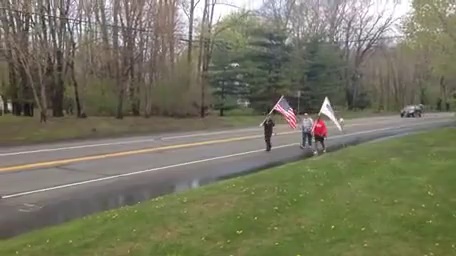} & \thumb{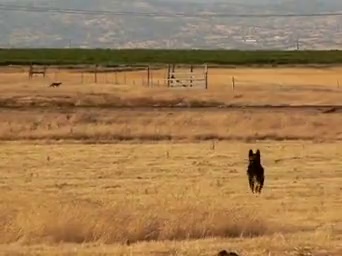} & \thumb{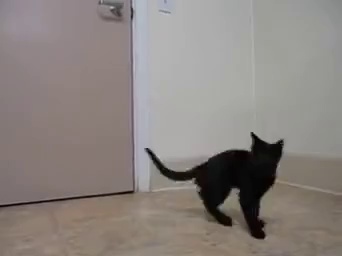} & \thumb{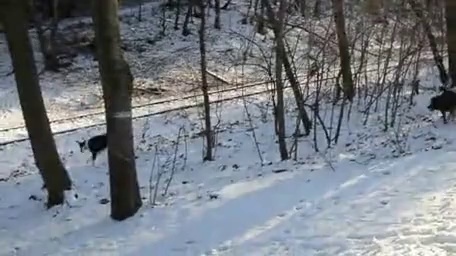} & \thumb{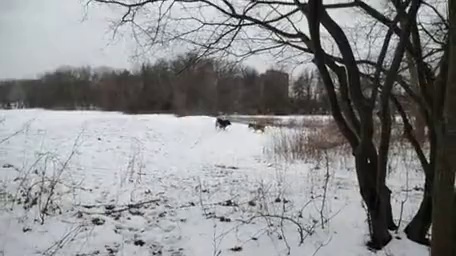} & \thumb{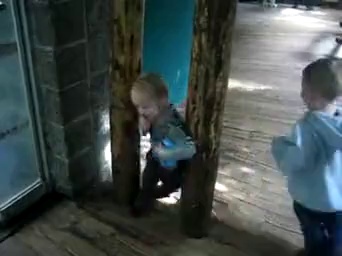} & \thumb{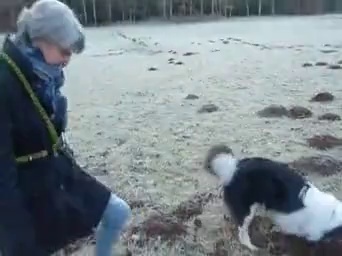} & \thumb{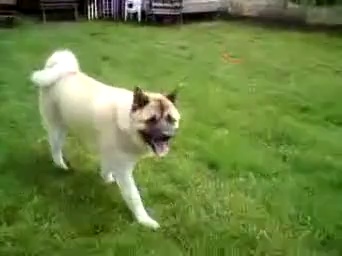} & \thumb{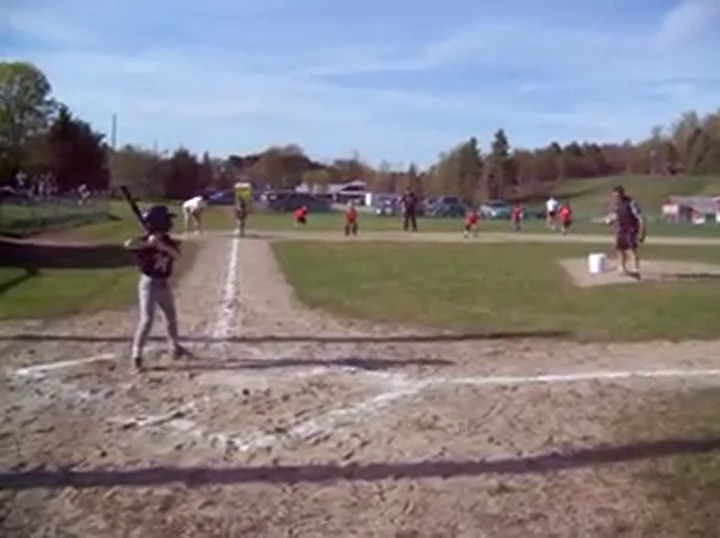} & \thumb{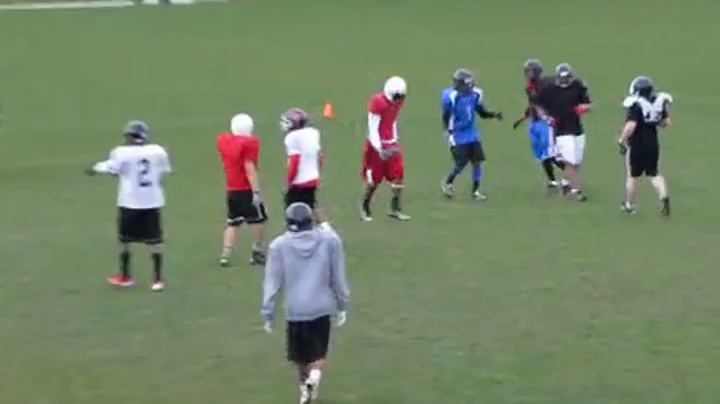} \\
\hline
\thumb{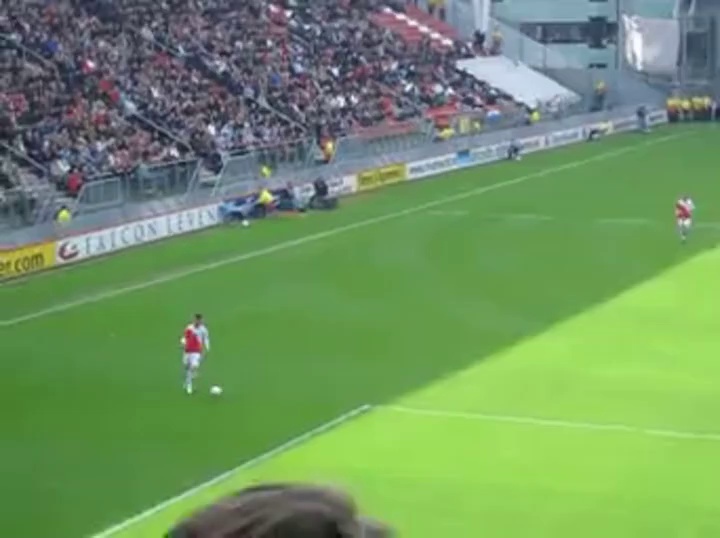} & \thumb{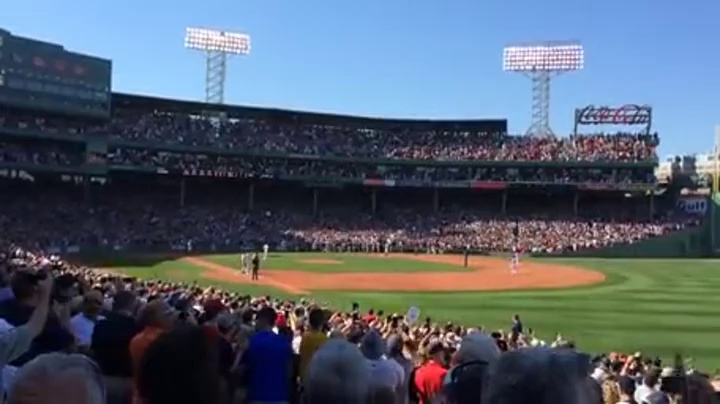} & \thumb{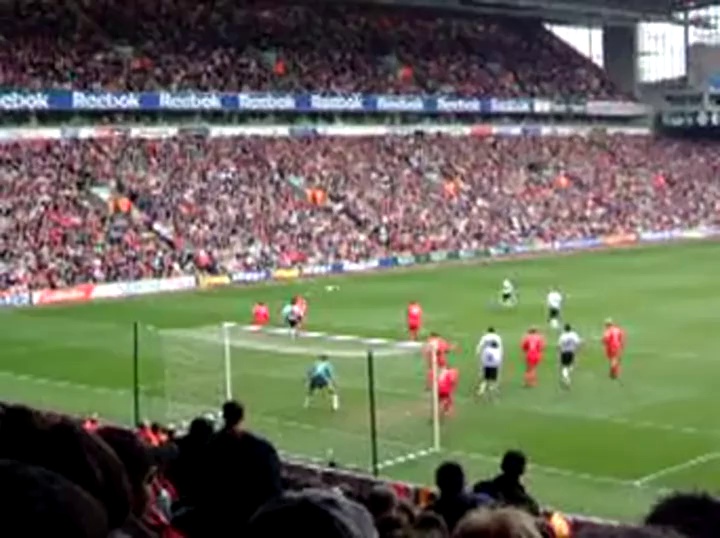} & \thumb{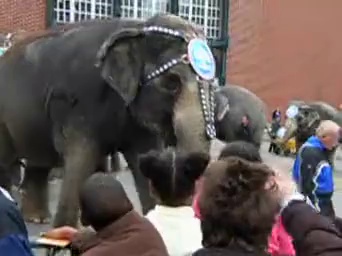} & \thumb{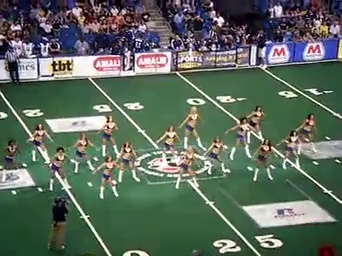} & \thumb{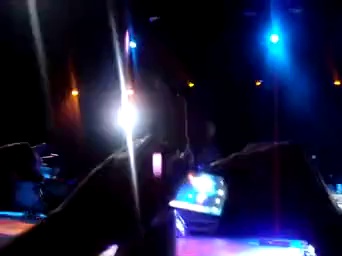} & \thumb{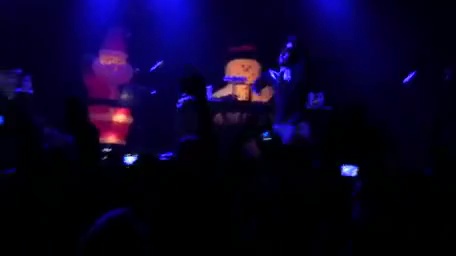} & \thumb{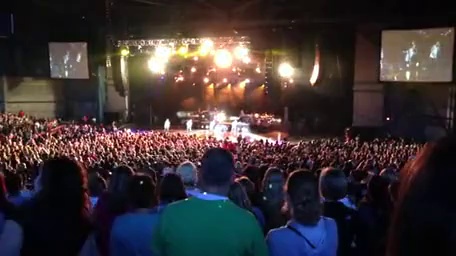} & \thumb{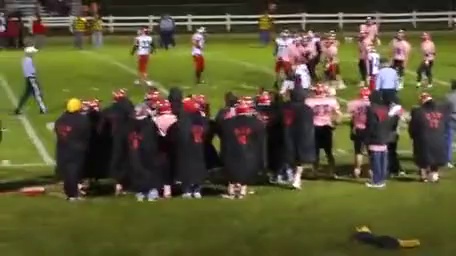} & \thumb{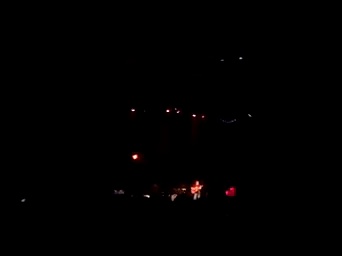} \\
\hline
\thumb{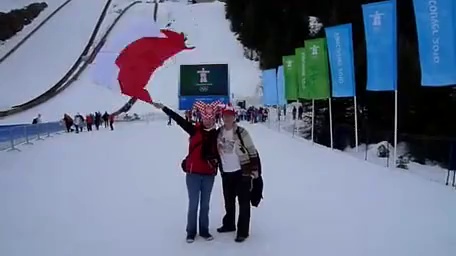} & \thumb{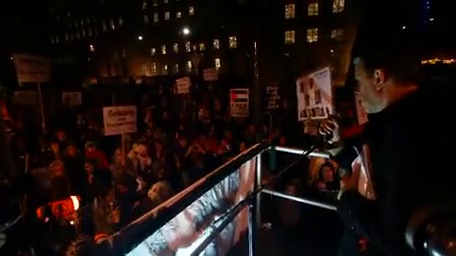} & \thumb{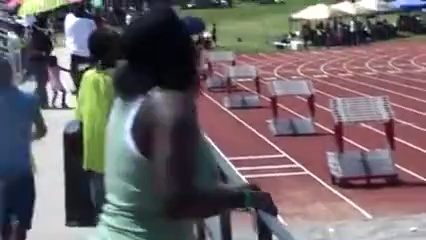} & \thumb{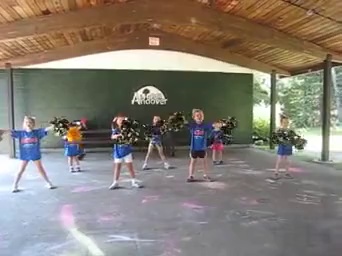} & \thumb{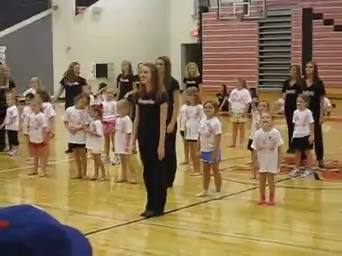} & \thumb{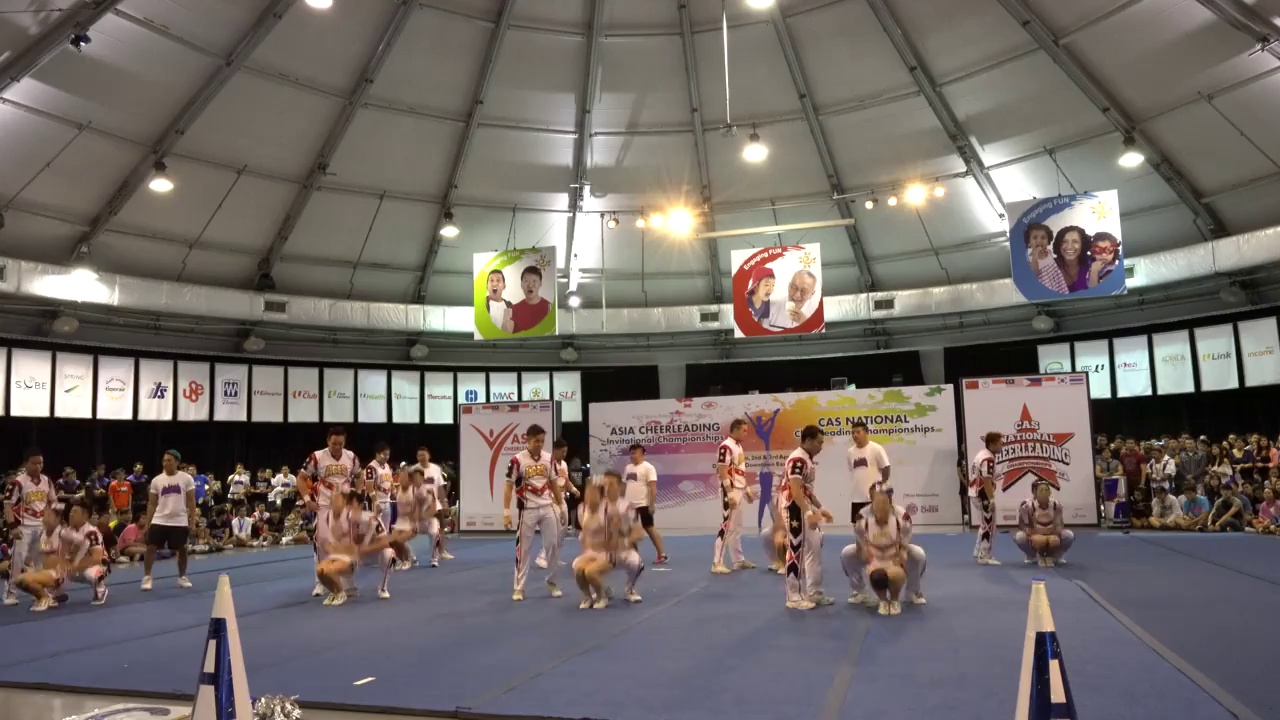} & \thumb{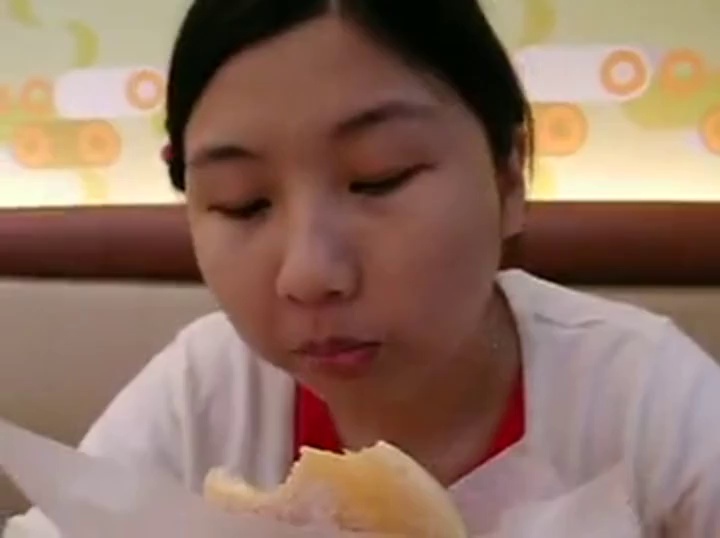} & \thumb{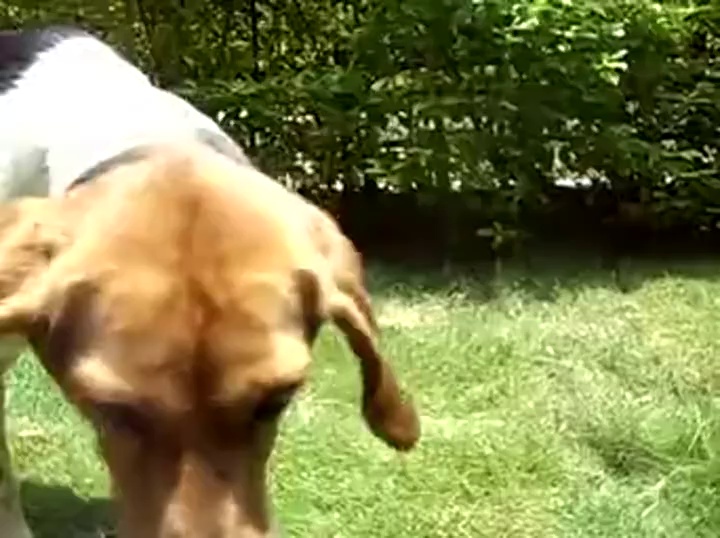} & \thumb{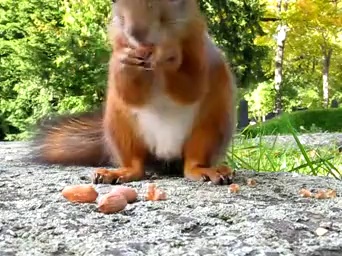} & \thumb{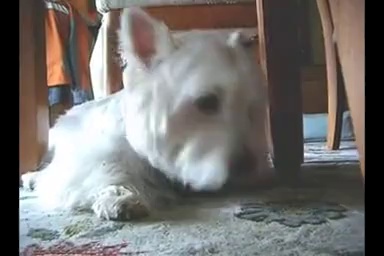} \\
\hline
\thumb{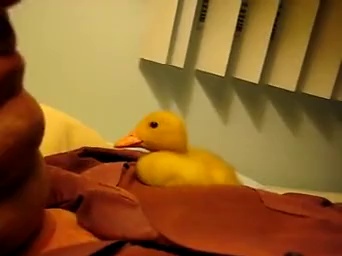} & \thumb{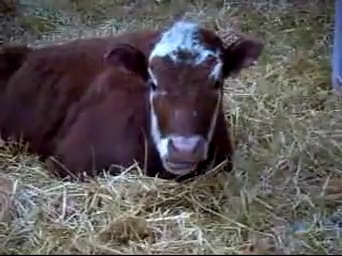} & \thumb{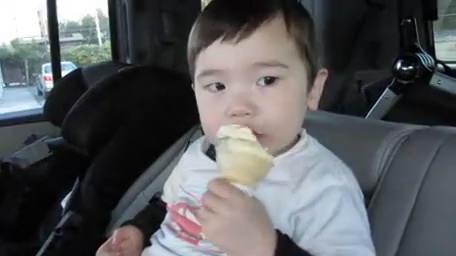} & \thumb{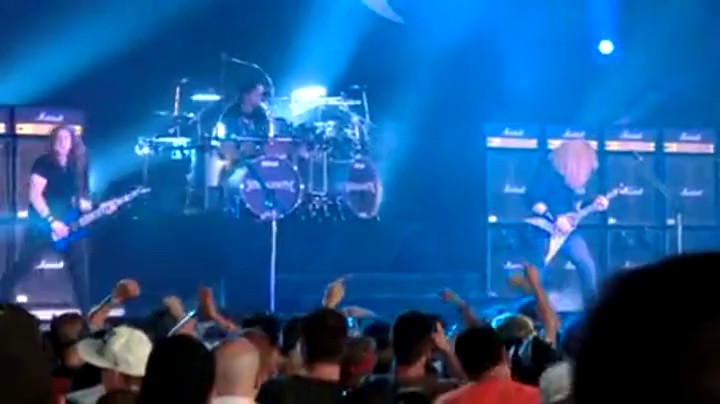} & \thumb{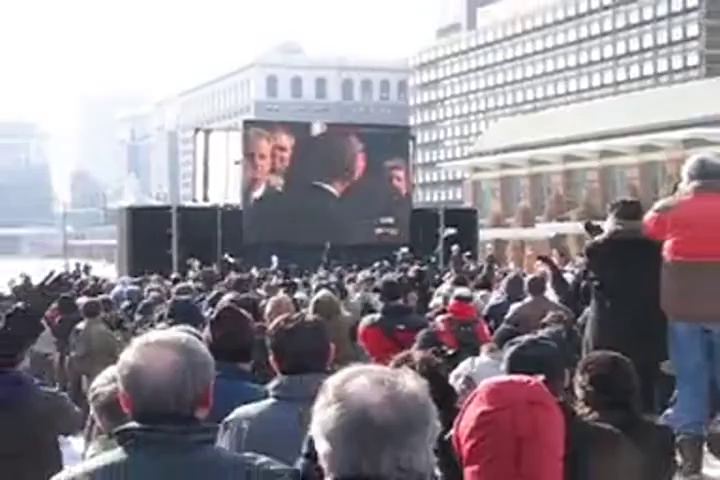} & \thumb{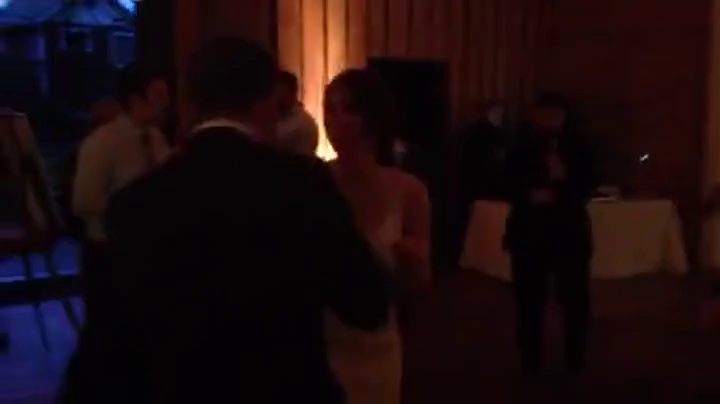} & \thumb{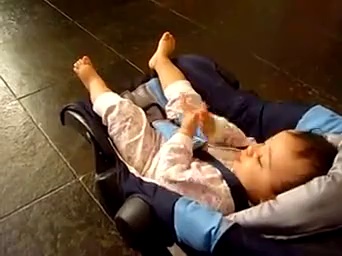} & \thumb{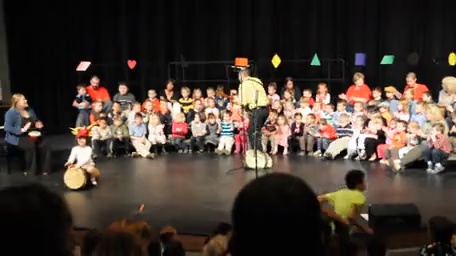} & \thumb{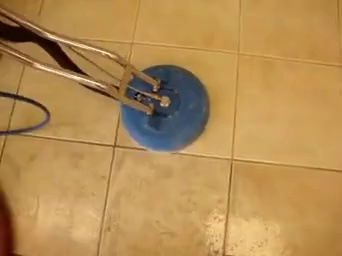} & \thumb{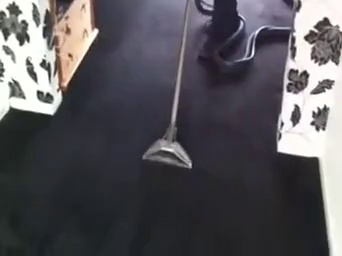} \\
\hline
\thumb{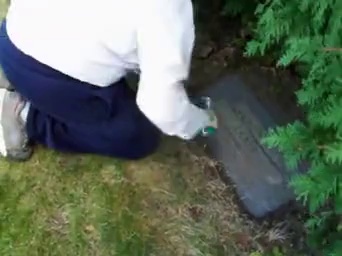} & \thumb{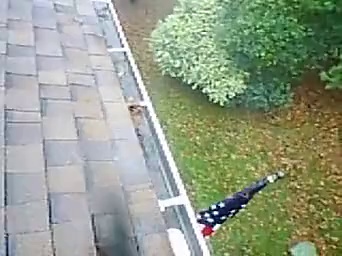} & \thumb{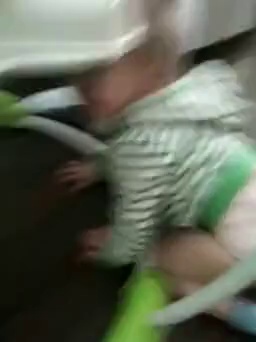} & \thumb{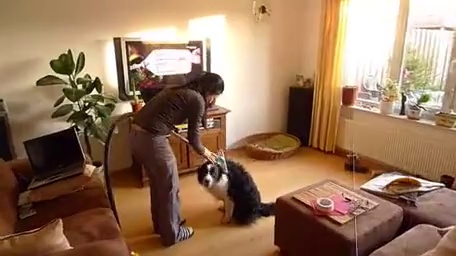} & \thumb{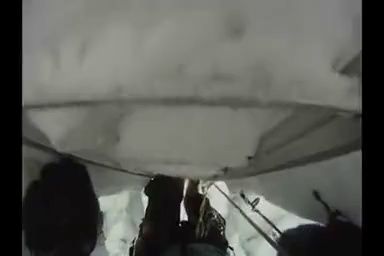} & \thumb{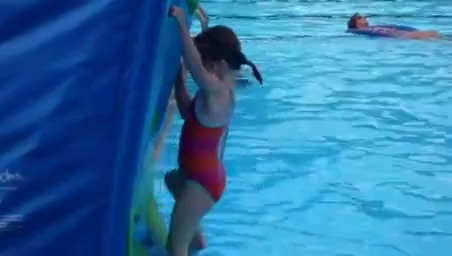} & \thumb{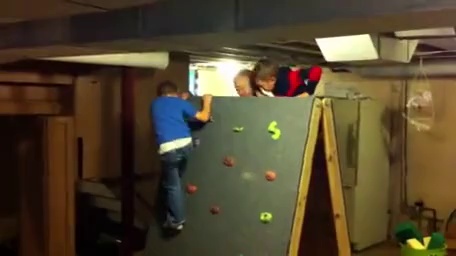} & \thumb{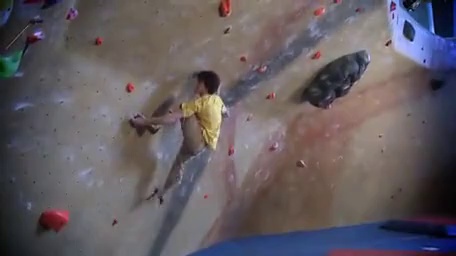} & \thumb{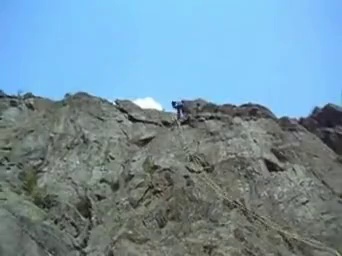} & \thumb{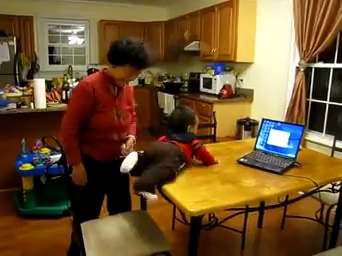} \\
\hline
\thumb{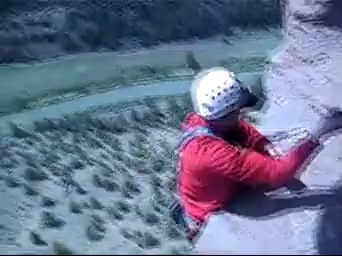} & \thumb{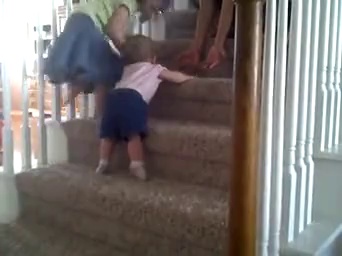} & \thumb{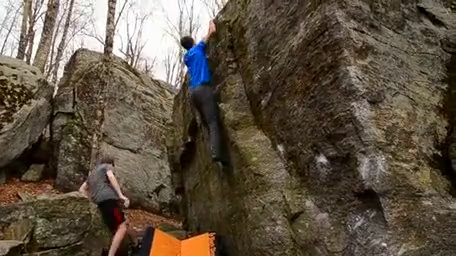} & \thumb{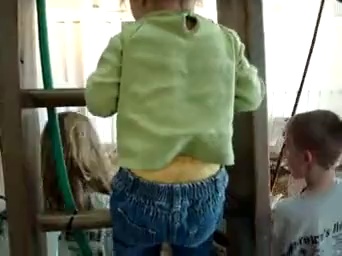} & \thumb{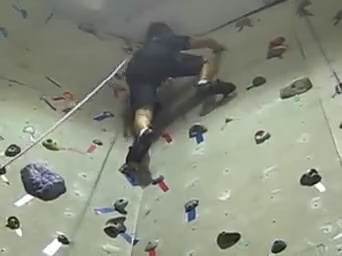} & \thumb{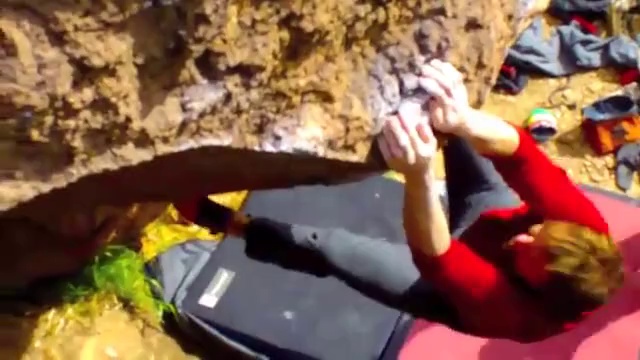} & \thumb{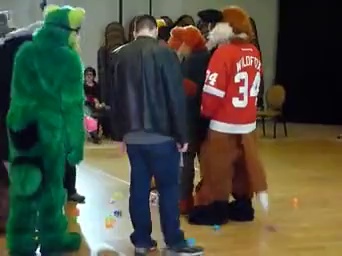} & \thumb{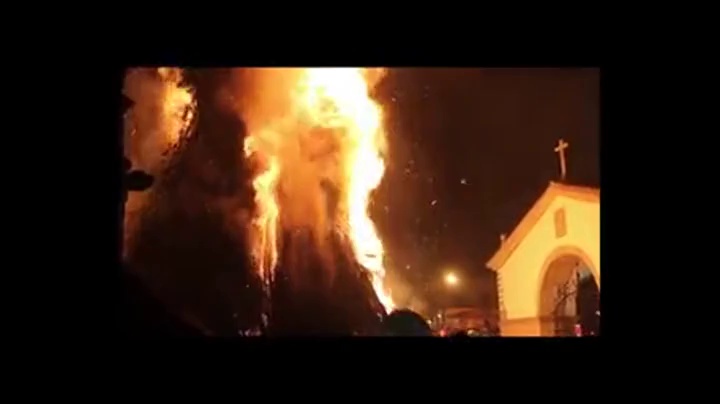} & \thumb{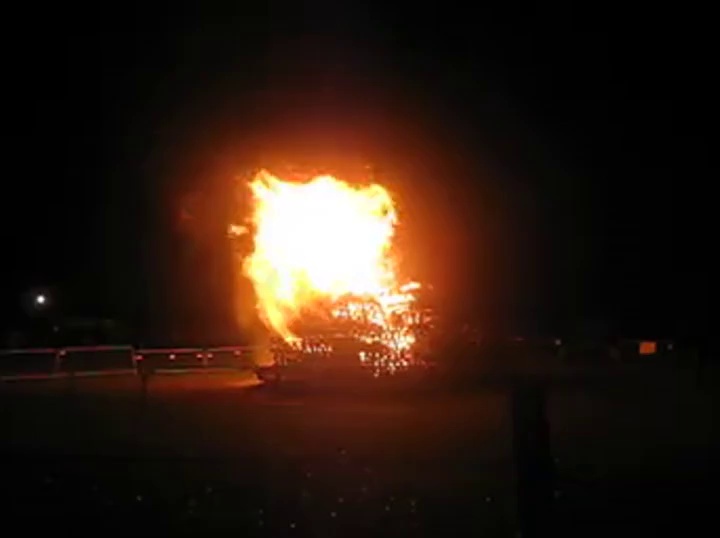} & \thumb{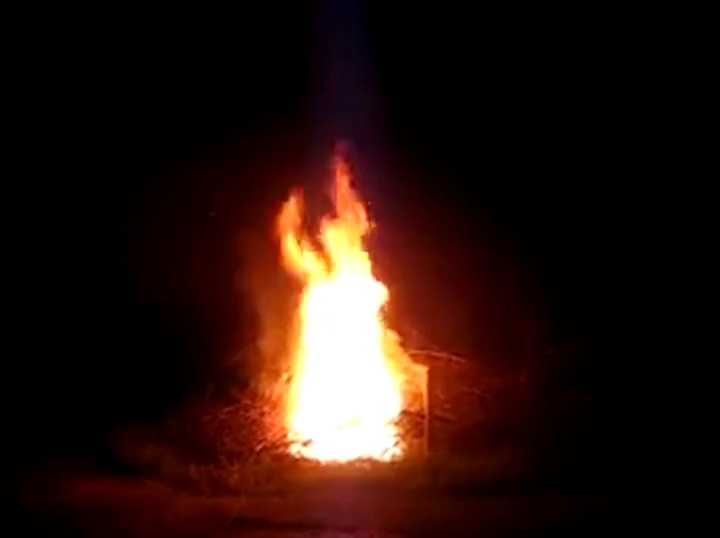} \\
\hline
\thumb{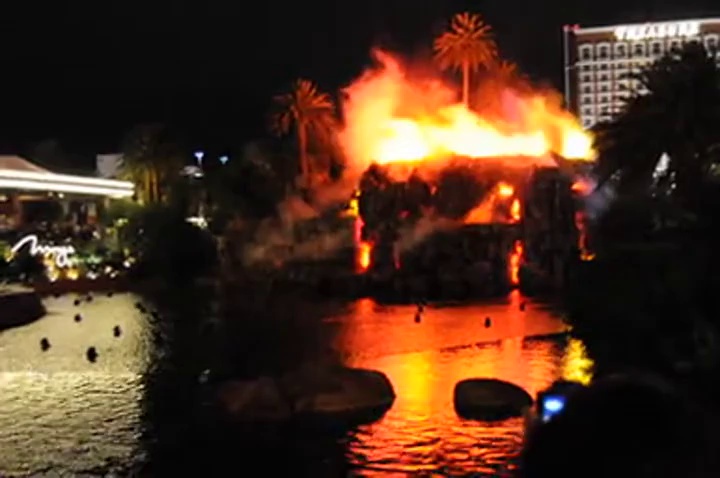} & \thumb{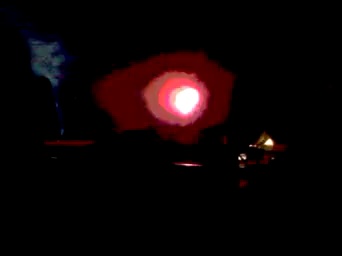} & \thumb{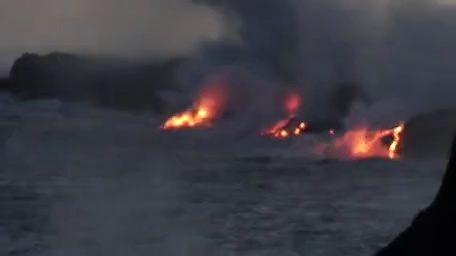} & \thumb{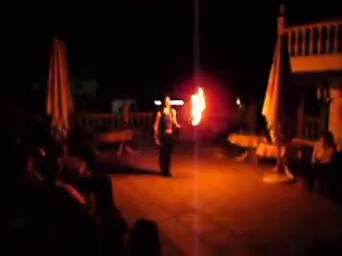} & \thumb{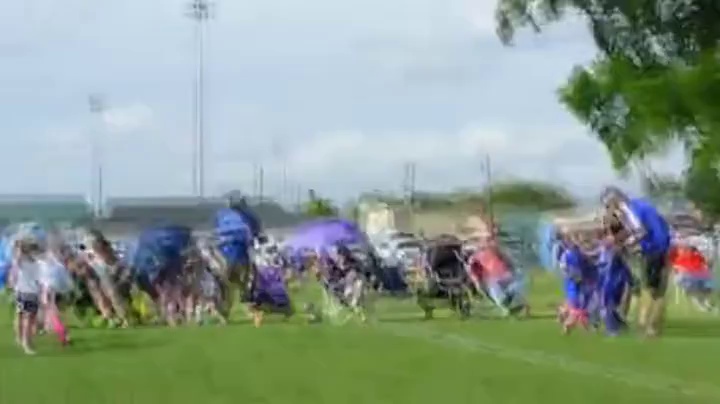} & \thumb{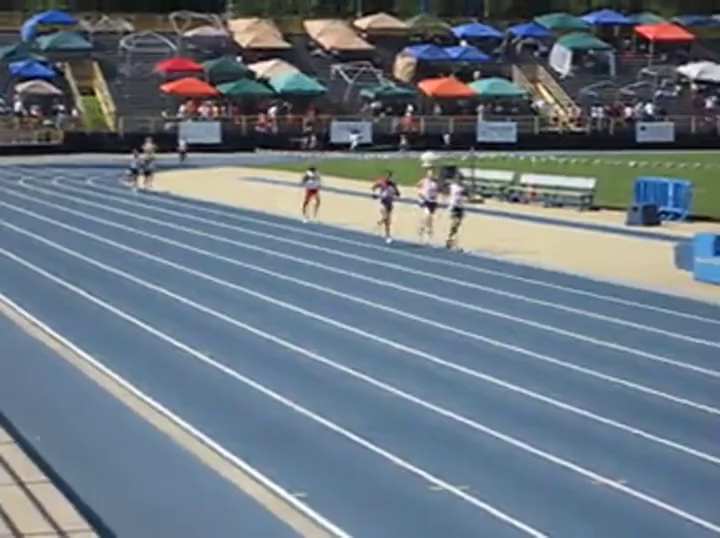} & \thumb{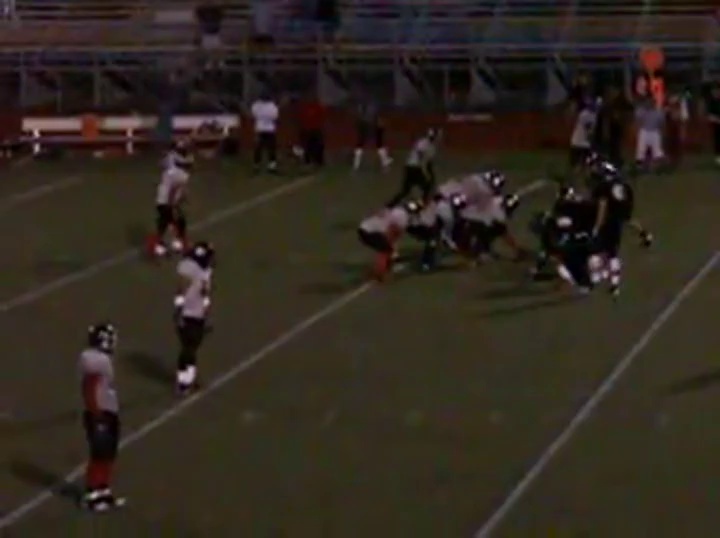} & \thumb{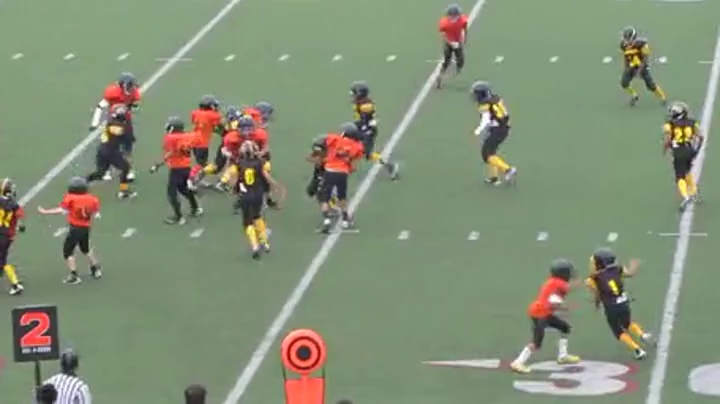} & \thumb{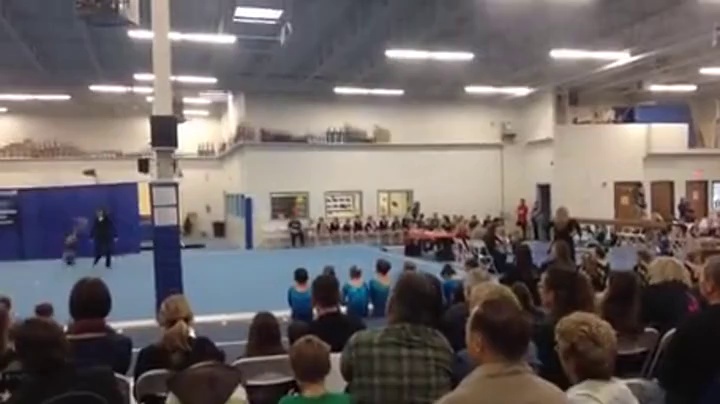} & \thumb{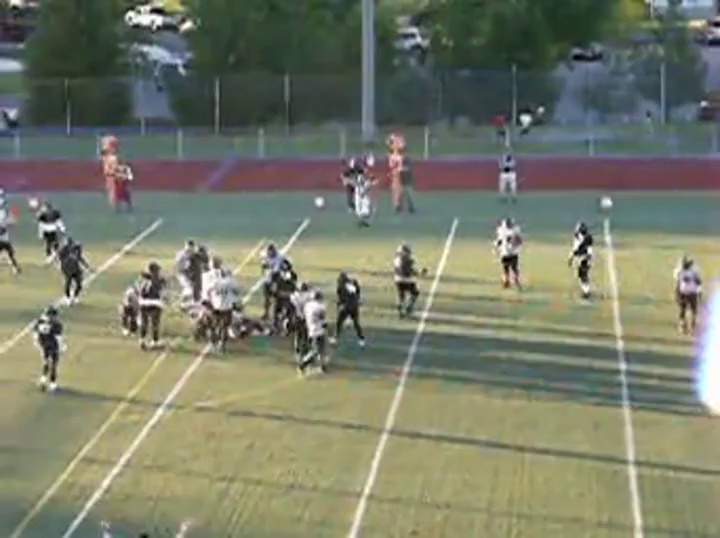} \\
\hline
\thumb{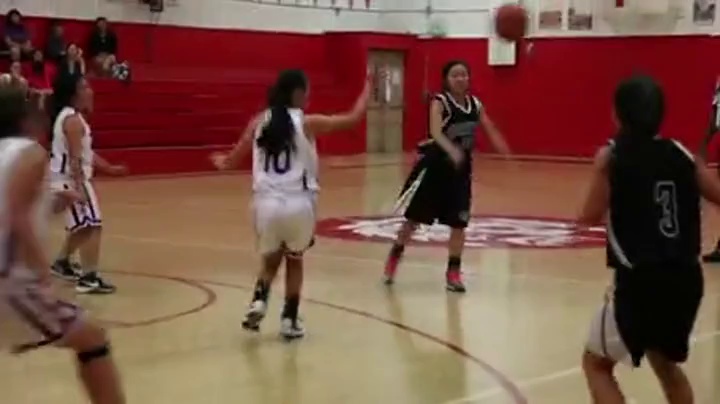} & \thumb{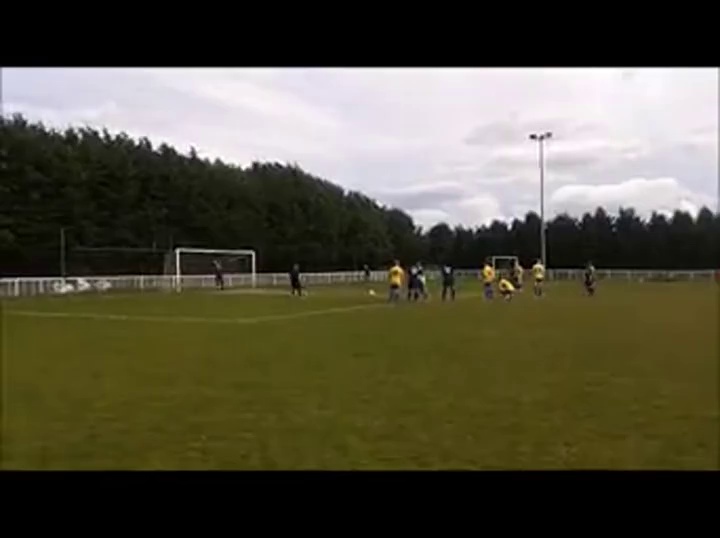} & \thumb{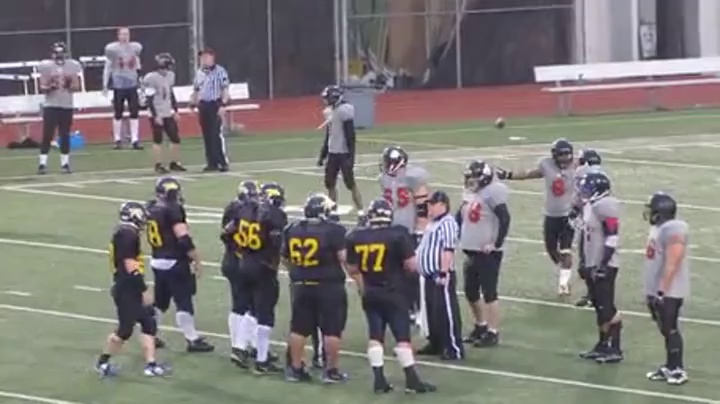} & \thumb{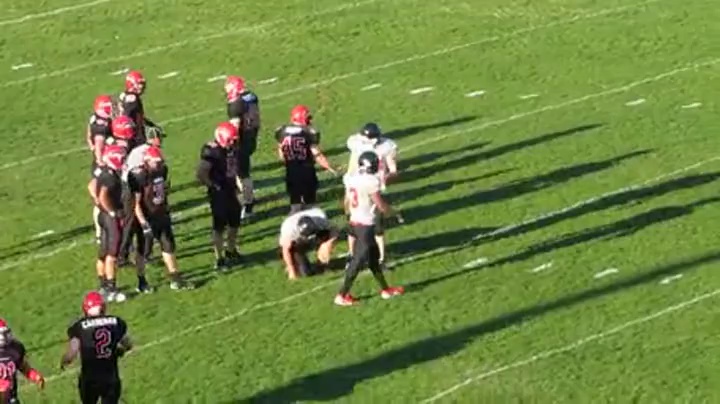} & \thumb{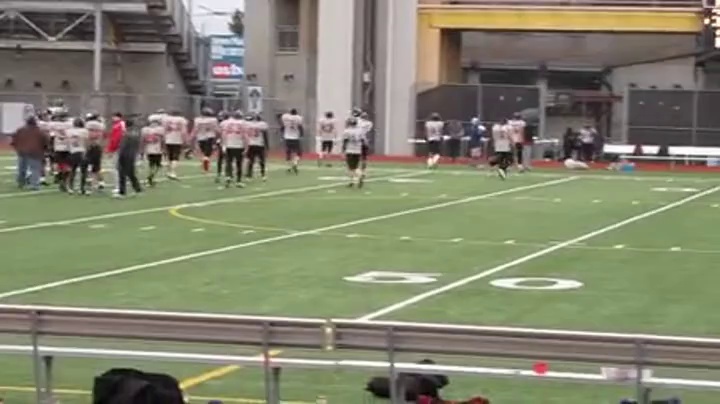} & \thumb{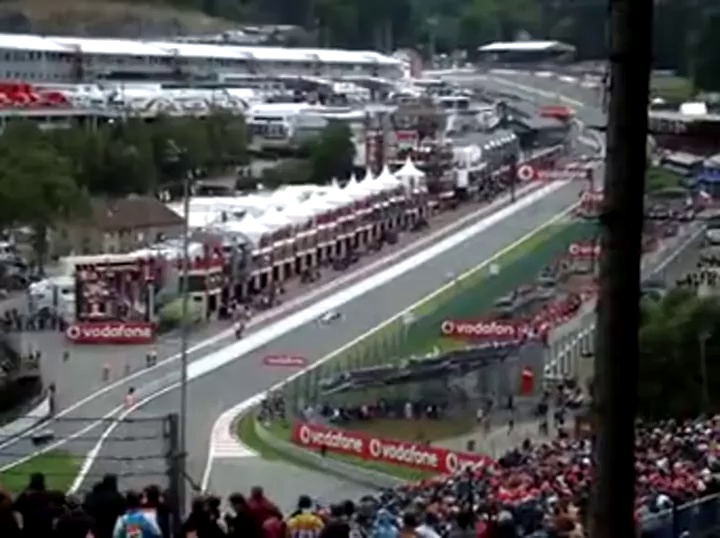} & \thumb{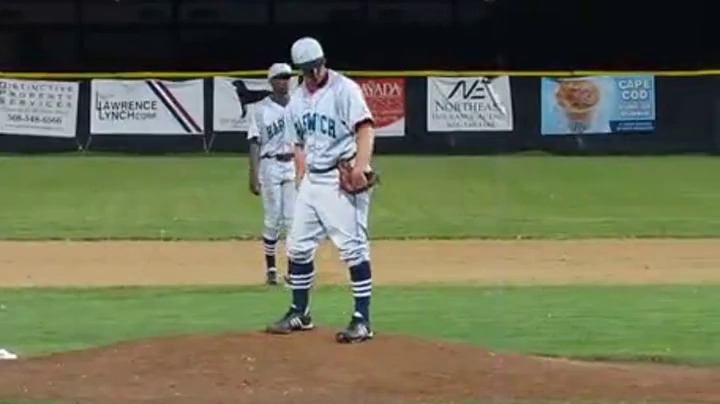} & \thumb{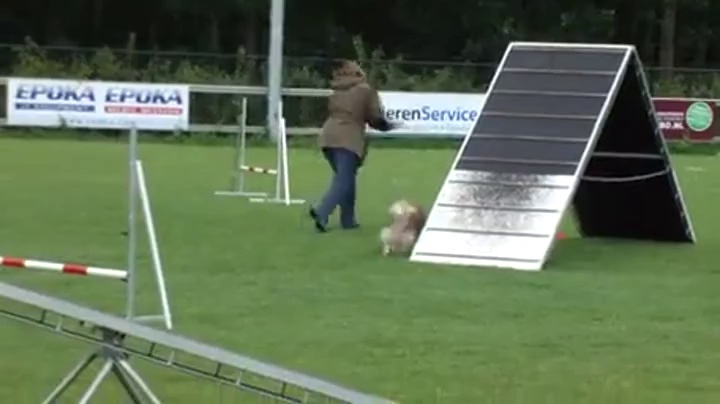} & \thumb{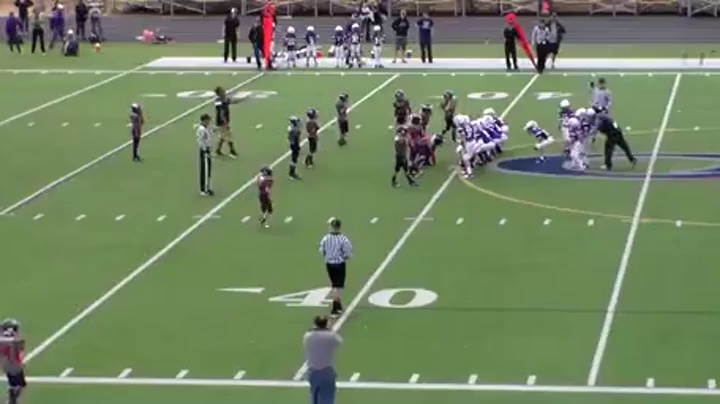} & \thumb{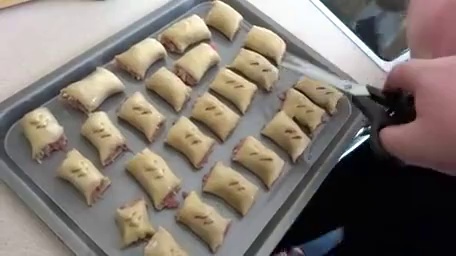} \\
\hline
\end{tabular}
\label{tab:thumbnails}
\end{table*}

\subsection{Using M3DDM for Outpainting}

M3DDM (Hierarchical Masked 3D Diffusion Model)~\cite{fan2024hierarchical}  offers an option to choose the aspect ratio  for the  output video and we chose a 1:1 ratio considering the importance of focus in the memorability of  video content.
M3DDM uses  masking techniques and global-frame features to achieve  outpainting. During inference, the model can use guide frames (first and last raw frames) to improve the quality and temporal consistency of the generated video. These  components define the M3DDM approach for video outpainting, focusing on maintaining temporal consistency and reducing artefacts.

\begin{figure}[ht]
\centering
\includegraphics[width=0.8\linewidth]{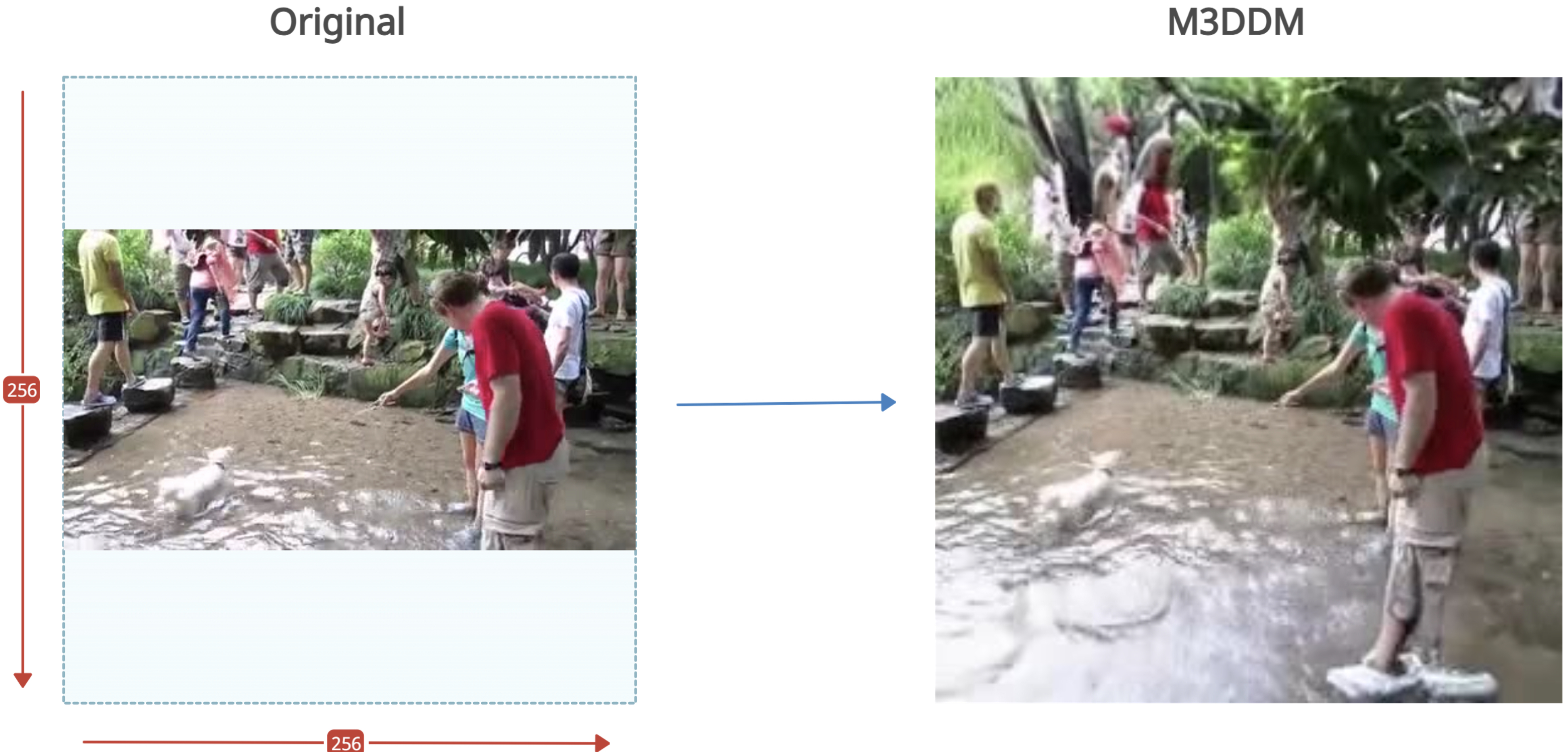}
\caption{Sample outpainting of a video frame using M3DDM, which allows for only 256x256 (1:1) resolution for outpainted videos.
\label{fig:m3ddm256}}
\end{figure}

For each of our 100 videos M3DDM  was used to generate outpainted versions by filling in  content outside the original frames. For example, if the original video was a landscape image of size 1280 x 720 (720p),  outpainting would infill a mask at the top and bottom namely an additional 1280x280 at the top and at the bottom and the generated video would be downsampled within the model to 266 x 256 pixels. 
Figure~\ref{fig:m3ddm256} shows an example original frame and outpainted using M3DDM.

\subsection{Using MOTIA for Outpainting}

The MOTIA (Mastering Video Outpainting Through Input-Specific Adaptation)~\cite{wang2024your} model  leverages intrinsic data-specific patterns of the source video and image/video generative priors for outpainting. The model also accepts a text parameter of either keywords of the objects in the frame or the description of the video itself.
We used the description of each video provided in the Memento10k dataset consisting of text transcriptions of the subjects, actions and surroundings in each video. MOTIA involves pseudo outpainting learning on the source video to capture essential patterns. Random masks are then added to the video, and the model learns to fill these masked regions. Those learned patterns are then generalised to produce outpainting  for the entire video.

\begin{figure}[htb]
\centering
\includegraphics[width=0.8\linewidth]{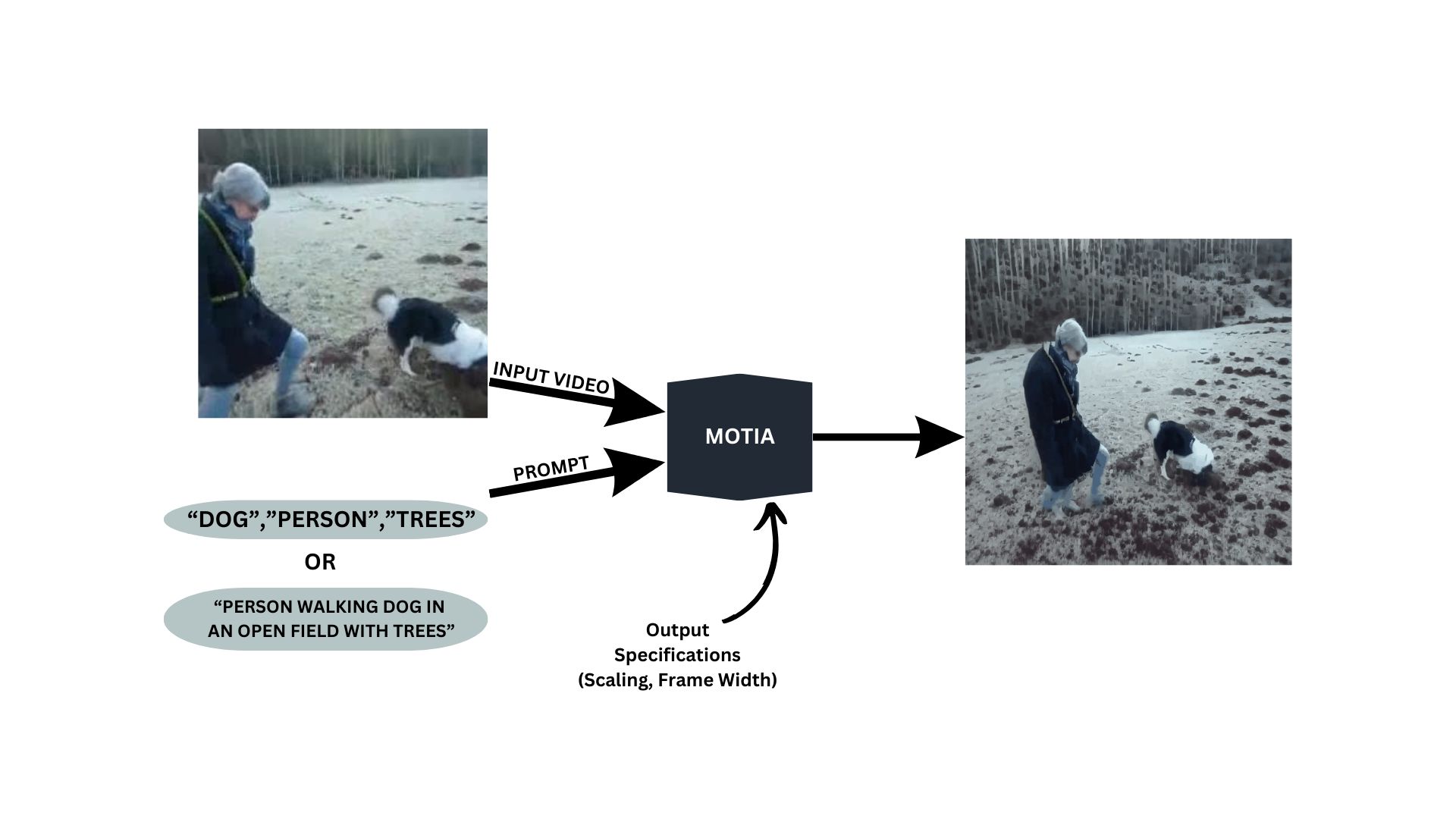}
\caption{A breakdown of the MOTIA approach to video outpainting.
\label{fig:motia_bre}}
\end{figure}

Unlike M3DDM, the MOTIA model offers flexibility in the size of the generated video frames and the position of the original frame within the  generated output. We   used a 1:1 aspect ratio for consistency in our evaluation of both  M3DDM and MOTIA. Figure~\ref{fig:motia_bre} shows a simplified breakdown of the MOTIA process and refers to some of the other parameters such as scaling and frame width. This highlights the flexibility of MOTIA compared to M3DDM.

\subsection{Predicting Video Memorability and Evaluation}

To compute the memorability of an original and of  outpainted videos, we used a vision transformer model fine-tuned specifically for video memorability prediction used in~\cite{cummins2022analysing}, with higher scores indicating higher memorability.
Each video, both original and outpainted, was fed into the vision transformer model to obtain memorability scores.
The primary metric used for evaluation of the effect of outpainting is the change in those memorability scores for each video.  

\begin{figure}[htb]
\centering
\includegraphics[width=\linewidth]{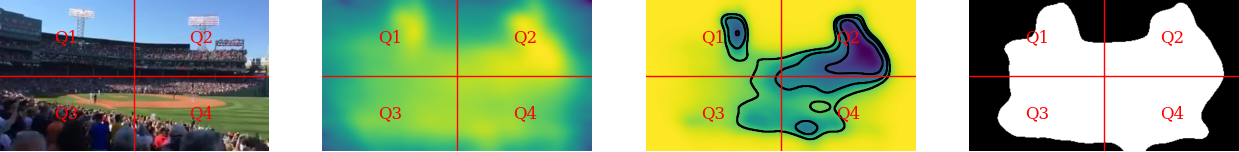}
\caption{Example saliency map and four quadrants.
\label{fig:quadrants}}
\end{figure}

In addition to outpainting each original video we also analysed the effect that saliency-based outpainting can have on  memorability scores. Saliency is a well-established technique in computer vision and refers to the visual prominence or attention-drawing nature of aspects within an image~\cite{ullah2020brief}. We generated a saliency map for each frame and divideded it into four quadrants as shown in Figure~\ref{fig:quadrants} to determine which quadrant contained the most salient part of the video frame. We then recognised the elements (e.g. objects) present in the most salient quadrant and used keywords for these as part of the input to MOTIA outpainting. We then focus the outpainting around this most salient quadrant. To illustrate, in the example in Figure~\ref{fig:quadrants} we  outpainted around quadrant 2, so adding only to the top and right side of the frame. 

\section{Experimental Results}

Our experimental procedure involved generating outpainted videos using the M3DDM and MOTIA models, computing memorability scores for  original and outpainted videos using the vision transformer model from~\cite{sweeney2021predicting} then comparing changes in memorability scores to assess the impact of outpainting.
The memorability scores for the 100 original videos that we conducted the experiments on ranged from 0.48 to 0.95.

\begin{figure}[ht]
\centering
\includegraphics[width=0.8\linewidth]{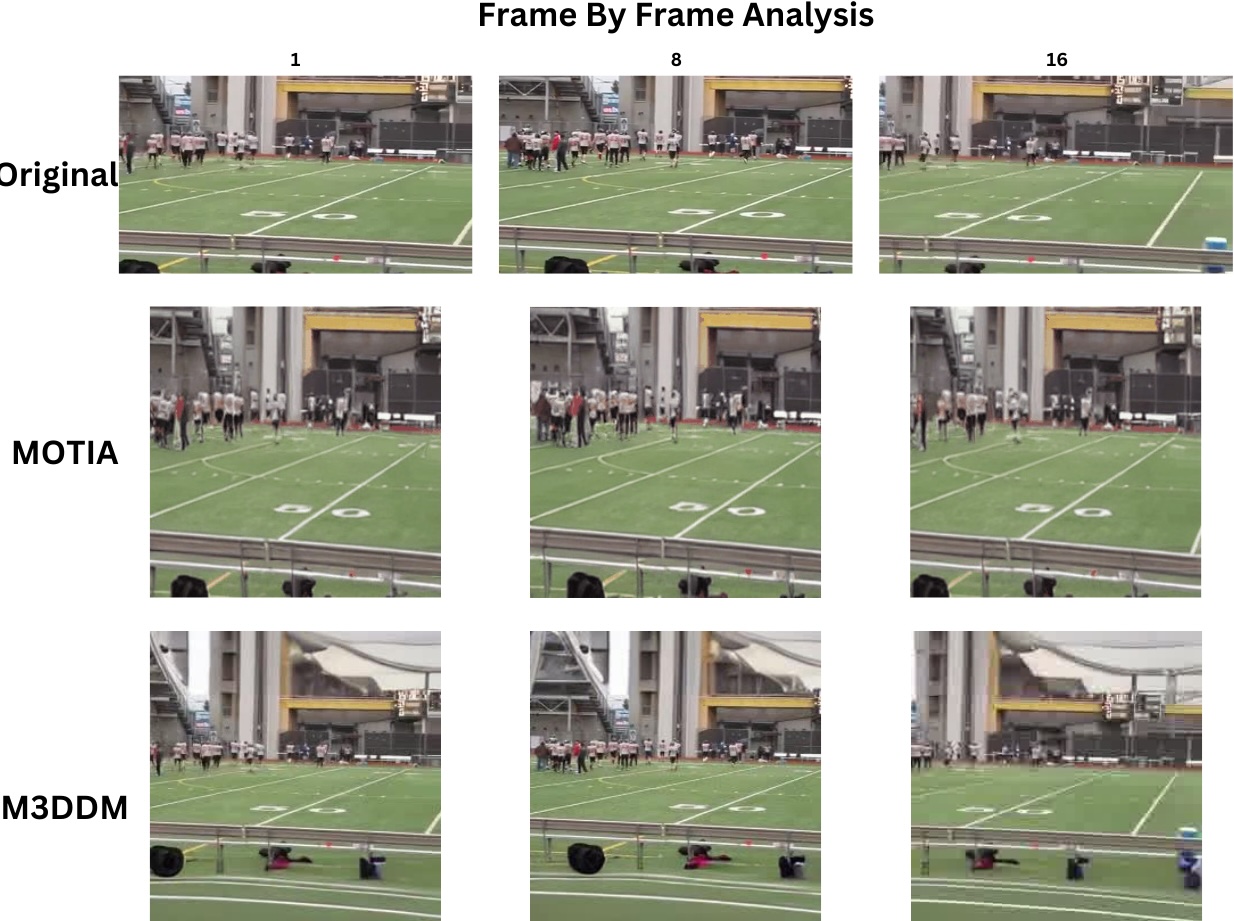}
\caption{Original video frames with outpainted results for MOTIA and M3DDM models.}
\label{fig:comparison}
\end{figure}

Figure~\ref{fig:comparison} illustrates the results of outpainting three sample frames from a video, 
the 1st, 8th and 16th. Memorability scores are shown in Table~\ref{tab:score_changes} indicating some improvements but it should be noted that the models appear to perform well in some frames and not well in others.

\begin{table}[htb]
    \centering
        \caption{Changes in memorability scores for 3 frames in Figure~\ref{fig:comparison} from outpainting.}
    \begin{tabular}{lccc}
    \toprule
         & ~~~~Frame 1~~~~ & ~~~~Frame 8~~~~ & ~~~~Frame 16~~~~ \\
         \midrule
       Original Frame~~~~  & 0.668 & 0.648 & 0.631 \\
       MOTIA  & 0.675 & 0.672  & 0.626 \\
       M3DDM & 0.687 & 0.691 & 0.622 \\
       \bottomrule
    \end{tabular}
    \label{tab:score_changes}
\end{table}

\begin{figure}[htb]
\centering
\includegraphics[width=\linewidth]{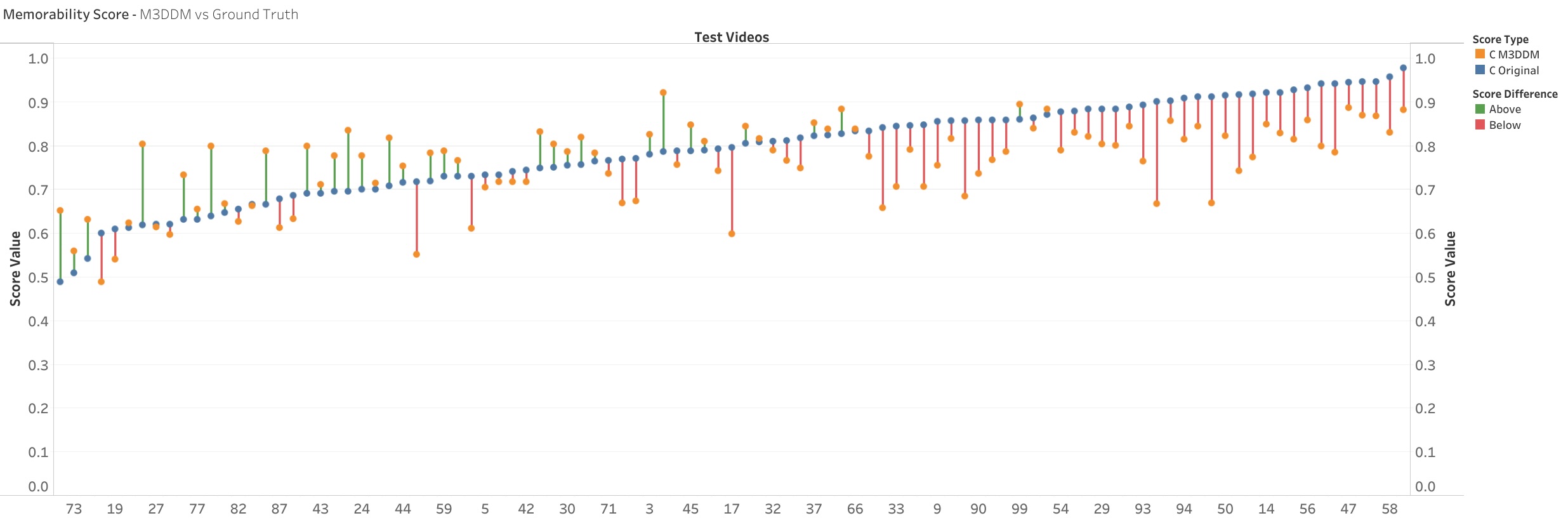}
\caption{Changes in memorability scores for test videos when using M3DDM for outpainting where average memorability score is 0.759. The x-axis refers to video numbers from within the Memento10k collection.}
\label{fig:db1}
\end{figure}

\begin{figure}[htb]
\centering
\includegraphics[width=\linewidth]{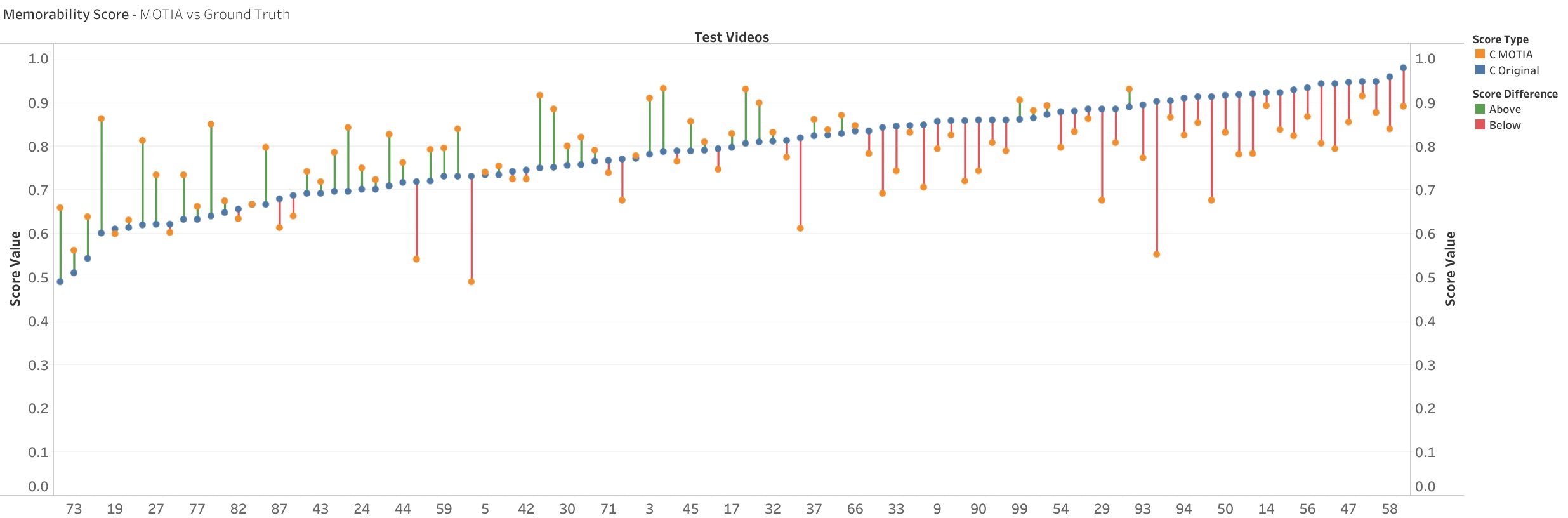}
\caption{Changes in memorability scores for test videos when using MOTIA for outpainting where average memorability score is 0.777. The x-axis refers to video numbers from within the Memento10k collection.}
\label{fig:db2}
\end{figure}

If we examine the dumbbell charts shown in Figures~\ref{fig:db1} and \ref{fig:db2}, these show the increases and decreases in memorability scores for the 100 test videos. Blue dots identify the ground truth scores while  orange dots are   scores for the outpainted videos. Green lines indicate an improvement in  memorability scores while red lines indicate a decrease. 

\begin{figure}[h]
\centering
\includegraphics[width=0.7\linewidth]{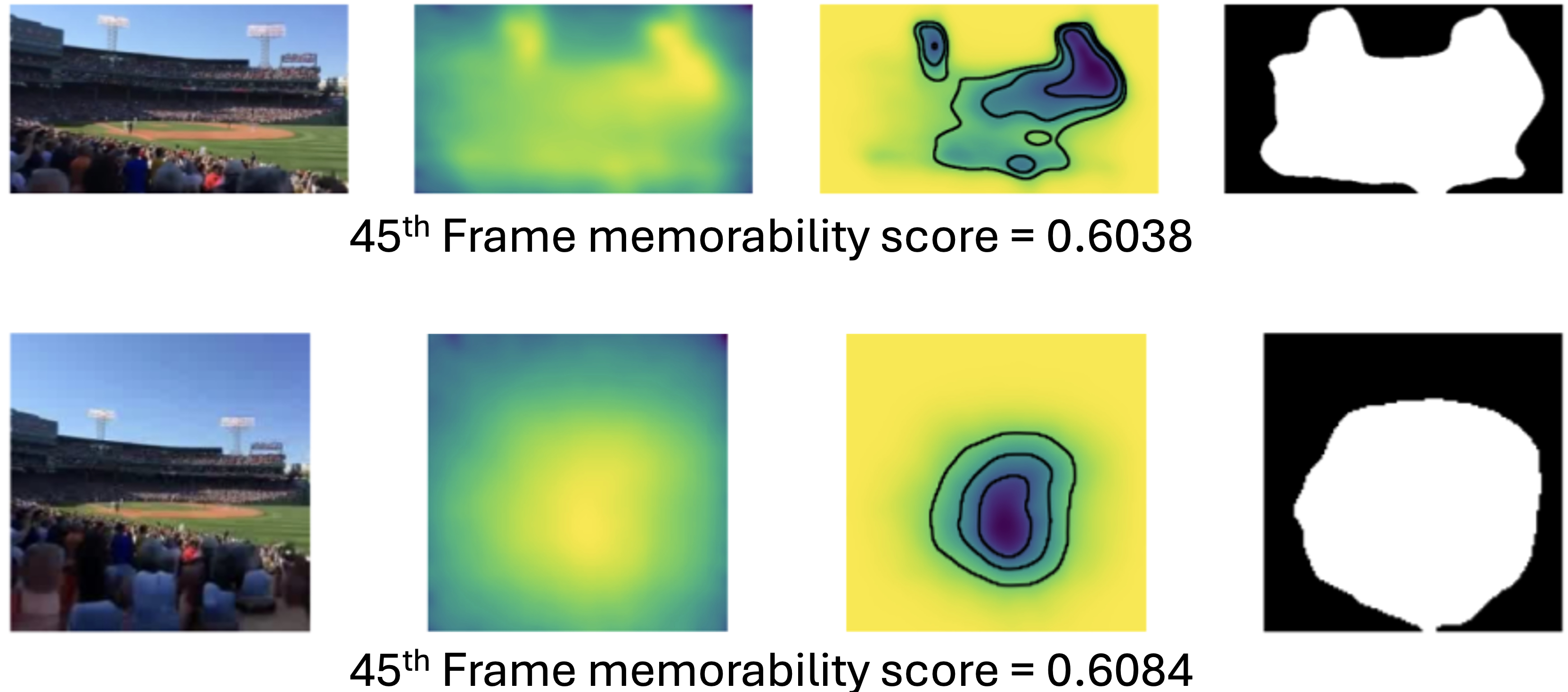}
\caption{Saliency mapping of a frame before and after outpainting with M3DDM. Top row shows the original video, its saliency heatmap, distribution of salient parts and the binary threshold.\label{fig:saliency}}
\end{figure}

We also carried out experiments with saliency-based outpainting by using saliency to determine the best areas of frames for outpainting as additional input into the MOTIA model. Earlier, Figure~\ref{fig:quadrants} showed an example of the most salient quadrant of an image and in Figure~\ref{fig:saliency} we show how we use quadrant 2 from the original image to prompt the MOTIA model with keywords like ``Stadium, Floodlights, Stands". The first image in the top row shows the original frame from a video of a baseball game with the stadium, spectators, field, and sky all visible. The second image in that first row shows the saliency map  before applying M3DDM while the third image  highlights the distribution of the salient parts of the original image. The last image is a binary threshold of  salient areas. In the second row we show the same images but for the  frame outpainted with M2DDM, and its salient areas and distribution. In this second row we see more sky has been added to the top and more infilling of spectators has been added to the bottom. he third image shows a more concentrated and defined area of high saliency, indicating a shift in visual attention to specific parts of the frame. The memorability scores for the 45th frame reveal an increase from 0.6038 before M3DDM to 0.6084 after its application. 

While this did  change the memorability scores, the effect was not uniform across the test videos, meaning it did not yield any significant difference in the memorability scores as compared to using MOTIA without saliency. 

From these visualisations we are able to identify two important points. Firstly, for both models, for videos with low initial memorability there is an increase in memorability scores as a result of  outpainting. Secondly, the mean memorability score for videos outpainted with MOTIA showed a marginal improvement, while  videos outpainted with M3DDM show a drop in average memorability score. Regarding the first point, an explanation is that when the borders are extended on videos with high initial scores, the areas of saliency weaken or there is a larger screensize of content to focus on. The second finding indicates that overall MOTIA performs better that M3DMM for outpainting due to a multitude of possible reasons, including the assistance of text-based input-specific adaptation.


\section{Conclusions}

The ide of actively manipulating videos by outpainting and also gathering information on the impact of outpainting based on image saliency on video memorability scores is relatively new.  
In this paper we examined the use of video outpainting to enhance short-form video memorability using    two advanced generative models, the Masked 3D Diffusion Model (M3DDM) and Mastering Video Outpainting Through Input-Specific Adaptation (MOTIA). 
Our experiments revealed several  findings regarding the impact of outpainting. First, we observed that outpainting generally improved  memorability of videos that  had  low memorability but for videos with high original memorability, outpainting tended to diminish this which was observed using both inpainting models we used. This suggests that outpainting can be  beneficial for enhancing the recall of less memorable content, but introduces distractions or reduces the focus on key elements in already memorable videos.
We also used saliency maps to analyse how changes in  video frames correlated with memorability scores and found a significant increase in memorability when salient parts of the image were brought into the center of the frame. However, this effect was not uniform across all videos, indicating that the benefits of outpainting are context-dependent.

Overall, while both M3DDM and MOTIA demonstrated effectiveness in improving video memorability through outpainting, MOTIA showed a marginally better performance but the significance of this is not clear and the results are inconsistent in terms of which videos are improved or not. This can be attributed to its innovative use of input-specific adaptation and pattern-aware outpainting techniques, which allowed for more contextually aware modifications to the videos but because of the close performances of the two models it challenging to draw a clear conclusion on which model performs better..
Our study highlights the potential of diffusion-based outpainting  to enhance video memorability, however, the computational demands of these models and the variable impact on different types of videos underscores the need for further research.

\begin{acknowledgments}

This work was partly-supported by Research Ireland under Grant Number: SFI/12/RC/2289\_P2, co-funded by the European Regional Development Fund.

\end{acknowledgments}

\bibliography{inpainting-paper.bib}

\begin{thebibliography}{23}
\expandafter\ifx\csname natexlab\endcsname\relax\def\natexlab#1{#1}\fi
\providecommand{\url}[1]{\texttt{#1}}
\providecommand{\href}[2]{#2}
\providecommand{\path}[1]{#1}
\providecommand{\DOIprefix}{doi:}
\providecommand{\ArXivprefix}{arXiv:}
\providecommand{\URLprefix}{URL: }
\providecommand{\Pubmedprefix}{pmid:}
\providecommand{\doi}[1]{\href{http://dx.doi.org/#1}{\path{#1}}}
\providecommand{\Pubmed}[1]{\href{pmid:#1}{\path{#1}}}
\providecommand{\bibinfo}[2]{#2}
\ifx\xfnm\relax \def\xfnm[#1]{\unskip,\space#1}\fi
\bibitem[{Cohendet et~al.(2019)Cohendet, Demarty, Duong, and Engilberge}]{Cohendet_2019_ICCV}
\bibinfo{author}{R.~Cohendet}, \bibinfo{author}{C.-H. Demarty}, \bibinfo{author}{N.~Q.~K. Duong}, \bibinfo{author}{M.~Engilberge},
\newblock \bibinfo{title}{Videomem: Constructing, analyzing, predicting short-term and long-term video memorability},
\newblock in: \bibinfo{booktitle}{Proceedings of the IEEE/CVF International Conference on Computer Vision (ICCV)}, \bibinfo{year}{2019}.
\bibitem[{Sweeney et~al.(2022)Sweeney, Constantin, Demarty, Fosco, de~Herrera, Halder, Healy, Ionescu, Matran-Fernandez, Smeaton et~al.}]{sweeney2022overview}
\bibinfo{author}{L.~Sweeney}, \bibinfo{author}{M.~G. Constantin}, \bibinfo{author}{C.-H. Demarty}, \bibinfo{author}{C.~Fosco}, \bibinfo{author}{A.~G.~S. de~Herrera}, \bibinfo{author}{S.~Halder}, \bibinfo{author}{G.~Healy}, \bibinfo{author}{B.~Ionescu}, \bibinfo{author}{A.~Matran-Fernandez}, \bibinfo{author}{A.~F. Smeaton}, et~al.,
\newblock \bibinfo{title}{Overview of the {MediaEval} 2022 predicting video memorability task},
\newblock \bibinfo{journal}{arXiv preprint arXiv:2212.06516}  (\bibinfo{year}{2022}).
\bibitem[{Sweeney et~al.(2020)Sweeney, Healy, and Smeaton}]{DBLP:journals/corr/abs-2012-15635}
\bibinfo{author}{L.~Sweeney}, \bibinfo{author}{G.~Healy}, \bibinfo{author}{A.~F. Smeaton},
\newblock \bibinfo{title}{Leveraging audio gestalt to predict media memorability},
\newblock \bibinfo{journal}{CoRR} \bibinfo{volume}{abs/2012.15635} (\bibinfo{year}{2020}). \URLprefix \url{https://arxiv.org/abs/2012.15635}. \href{http://arxiv.org/abs/2012.15635}{{\tt arXiv:2012.15635}}.
\bibitem[{Bainbridge et~al.(2017)Bainbridge, Dilks, and Oliva}]{BAINBRIDGE2017141}
\bibinfo{author}{W.~A. Bainbridge}, \bibinfo{author}{D.~D. Dilks}, \bibinfo{author}{A.~Oliva},
\newblock \bibinfo{title}{Memorability: A stimulus-driven perceptual neural signature distinctive from memory},
\newblock \bibinfo{journal}{NeuroImage} \bibinfo{volume}{149} (\bibinfo{year}{2017}) \bibinfo{pages}{141--152}. \URLprefix \url{https://www.sciencedirect.com/science/article/pii/S1053811917300861}. \DOIprefix\doi{https://doi.org/10.1016/j.neuroimage.2017.01.063}.
\bibitem[{Almog et~al.(2021)Almog, Naeini, Hu, Duerden, and Mohsenzadeh}]{10.31234/osf.io/kd29q}
\bibinfo{author}{G.~Almog}, \bibinfo{author}{S.~A. Naeini}, \bibinfo{author}{Y.~Hu}, \bibinfo{author}{E.~Duerden}, \bibinfo{author}{Y.~Mohsenzadeh}, \bibinfo{title}{Memoir dataset: Quantifying image memorability in adolescents}, \bibinfo{howpublished}{osf.io/preprints/psyarxiv/kd29q}, \bibinfo{year}{2021}.
\bibitem[{Shekar et~al.(2021)Shekar, Shetty, and Bhat}]{10.1007/978-981-16-1092-9_41}
\bibinfo{author}{B.~H. Shekar}, \bibinfo{author}{P.~R. Shetty}, \bibinfo{author}{S.~S. Bhat},
\newblock \bibinfo{title}{Complex gradient function based descriptor for iris biometrics and action recognition},
\newblock \bibinfo{journal}{Communications in Computer and Information Science}  (\bibinfo{year}{2021}) \bibinfo{pages}{489--501}. \DOIprefix\doi{10.1007/978-981-16-1092-9\_41}.
\bibitem[{Wang et~al.(2011)Wang, Konrad, Ishwar, Jing, and Rowley}]{wang2011image}
\bibinfo{author}{M.~Wang}, \bibinfo{author}{J.~Konrad}, \bibinfo{author}{P.~Ishwar}, \bibinfo{author}{K.~Jing}, \bibinfo{author}{H.~Rowley},
\newblock \bibinfo{title}{Image saliency: From intrinsic to extrinsic context},
\newblock in: \bibinfo{booktitle}{CVPR 2011}, \bibinfo{organization}{IEEE}, \bibinfo{year}{2011}, pp. \bibinfo{pages}{417--424}.
\bibitem[{Cummins et~al.(2022)Cummins, Sweeney, and Smeaton}]{cummins2022analysing}
\bibinfo{author}{S.~Cummins}, \bibinfo{author}{L.~Sweeney}, \bibinfo{author}{A.~F. Smeaton},
\newblock \bibinfo{title}{{Analysing the Memorability of a Procedural Crime-Drama TV Series, CSI}},
\newblock in: \bibinfo{booktitle}{Proceedings of the 19th International Conference on Content-based Multimedia Indexing}, \bibinfo{year}{2022}, pp. \bibinfo{pages}{174--180}.
\bibitem[{Guinaudeau and Xalabarder(2023)}]{guinaudeau2023textual}
\bibinfo{author}{C.~Guinaudeau}, \bibinfo{author}{A.~G. Xalabarder},
\newblock \bibinfo{title}{Textual analysis for video memorability prediction},
\newblock in: \bibinfo{booktitle}{Working Notes Proceedings of the MediaEval 2022 Workshop}, \bibinfo{year}{2023}.
\bibitem[{Hachchane et~al.(2020)Hachchane, Badri, Sahel, and Ruichek}]{10.11591/ijai.v9.i1.pp40-45}
\bibinfo{author}{I.~Hachchane}, \bibinfo{author}{A.~Badri}, \bibinfo{author}{A.~Sahel}, \bibinfo{author}{Y.~Ruichek},
\newblock \bibinfo{title}{Large-scale image-to-video face retrieval with convolutional neural network features},
\newblock \bibinfo{journal}{IAES International Journal of Artificial Intelligence (IJ-AI)} \bibinfo{volume}{9} (\bibinfo{year}{2020}) \bibinfo{pages}{40}. \DOIprefix\doi{10.11591/ijai.v9.i1.pp40-45}.
\bibitem[{Harini et~al.(2024)Harini, Singh, Kumar, Bhattacharyya, Baths, Chen, Shah, and Krishnamurthy}]{s2024longtermadmemorabilityunderstanding}
\bibinfo{author}{S.~Harini}, \bibinfo{author}{S.~Singh}, \bibinfo{author}{Y.~Kumar}, \bibinfo{author}{A.~Bhattacharyya}, \bibinfo{author}{V.~Baths}, \bibinfo{author}{C.~Chen}, \bibinfo{author}{R.~R. Shah}, \bibinfo{author}{B.~Krishnamurthy},
\newblock \bibinfo{title}{{Long-Term Ad Memorability: Understanding \& Generating Memorable Ads}},
\newblock \bibinfo{journal}{arXiv preprint arXiv:2309.00378}  (\bibinfo{year}{2024}).
\bibitem[{Newman et~al.(2020)Newman, Fosco, Casser, Lee, McNamara, and Oliva}]{newman2020multimodal}
\bibinfo{author}{A.~Newman}, \bibinfo{author}{C.~Fosco}, \bibinfo{author}{V.~Casser}, \bibinfo{author}{A.~Lee}, \bibinfo{author}{B.~McNamara}, \bibinfo{author}{A.~Oliva},
\newblock \bibinfo{title}{Multimodal memorability: Modeling effects of semantics and decay on video memorability},
\newblock in: \bibinfo{booktitle}{ECCV 2020: 16th European Conference on Computer Vision, Glasgow, UK, August 23--28, 2020, Proceedings, Part XVI 16}, \bibinfo{organization}{Springer}, \bibinfo{year}{2020}, pp. \bibinfo{pages}{223--240}.
\bibitem[{Sweeney et~al.(2021)Sweeney, Healy, and Smeaton}]{sweeney2021predicting}
\bibinfo{author}{L.~Sweeney}, \bibinfo{author}{G.~Healy}, \bibinfo{author}{A.~F. Smeaton},
\newblock \bibinfo{title}{Predicting media memorability: comparing visual, textual and auditory features},
\newblock \bibinfo{journal}{In: MediaEval 2021 Multimedia Benchmark, arXiv preprint arXiv:2112.07969}  (\bibinfo{year}{2021}).
\bibitem[{De~Herrera et~al.(2020)De~Herrera, Kiziltepe, Chamberlain, Constantin, Demarty, Doctor, Ionescu, and Smeaton}]{de2020overview}
\bibinfo{author}{A.~G.~S. De~Herrera}, \bibinfo{author}{R.~S. Kiziltepe}, \bibinfo{author}{J.~Chamberlain}, \bibinfo{author}{M.~G. Constantin}, \bibinfo{author}{C.-H. Demarty}, \bibinfo{author}{F.~Doctor}, \bibinfo{author}{B.~Ionescu}, \bibinfo{author}{A.~F. Smeaton},
\newblock \bibinfo{title}{{Overview of MediaEval 2020 predicting media memorability task: What makes a video memorable?}},
\newblock \bibinfo{journal}{arXiv preprint arXiv:2012.15650}  (\bibinfo{year}{2020}).
\bibitem[{Mudgal et~al.(2024)Mudgal, Wang, Sweeney, and Smeaton}]{mudgal2024using}
\bibinfo{author}{V.~Mudgal}, \bibinfo{author}{Q.~Wang}, \bibinfo{author}{L.~Sweeney}, \bibinfo{author}{A.~F. Smeaton},
\newblock \bibinfo{title}{Using saliency and cropping to improve video memorability},
\newblock in: \bibinfo{booktitle}{International Conference on Multimedia Modeling}, \bibinfo{organization}{Springer}, \bibinfo{year}{2024}, pp. \bibinfo{pages}{342--355}.
\bibitem[{Fan et~al.(2024)Fan, Guo, Gong, Wang, Ge, Jiang, Luo, and Zhan}]{fan2024hierarchical}
\bibinfo{author}{F.~Fan}, \bibinfo{author}{C.~Guo}, \bibinfo{author}{L.~Gong}, \bibinfo{author}{B.~Wang}, \bibinfo{author}{T.~Ge}, \bibinfo{author}{Y.~Jiang}, \bibinfo{author}{C.~Luo}, \bibinfo{author}{J.~Zhan},
\newblock \bibinfo{title}{Hierarchical masked 3d diffusion model for video outpainting},
\newblock \bibinfo{journal}{arXiv preprint arXiv:22309.02119}  (\bibinfo{year}{2024}).
\bibitem[{Dhariwal and Nichol(2021)}]{dhariwal2021diffusion}
\bibinfo{author}{P.~Dhariwal}, \bibinfo{author}{A.~Nichol},
\newblock \bibinfo{title}{Diffusion models beat gans on image synthesis},
\newblock \bibinfo{journal}{Advances in Neural Information Processing Systems} \bibinfo{volume}{34} (\bibinfo{year}{2021}) \bibinfo{pages}{8780--8794}.
\bibitem[{Tammineni et~al.(2024)Tammineni, Rayavarapu, Gottapu, and Goswami}]{10.35784/iapgos.5373}
\bibinfo{author}{S.~Tammineni}, \bibinfo{author}{S.~M. Rayavarapu}, \bibinfo{author}{S.~R. Gottapu}, \bibinfo{author}{R.~K. Goswami},
\newblock \bibinfo{title}{Digital image restoration using {SURF} algorithm},
\newblock \bibinfo{journal}{Informatyka, Automatyka, Pomiary W Gospodarce I Ochronie Środowiska} \bibinfo{volume}{14} (\bibinfo{year}{2024}) \bibinfo{pages}{37--40}. \DOIprefix\doi{10.35784/iapgos.5373}.
\bibitem[{Gao et~al.(2020)Gao, Saraf, Huang, and Kopf}]{gao2020flowedge}
\bibinfo{author}{C.~Gao}, \bibinfo{author}{A.~Saraf}, \bibinfo{author}{J.-B. Huang}, \bibinfo{author}{J.~Kopf},
\newblock \bibinfo{title}{Flow-edge guided video completion},
\newblock \bibinfo{journal}{arXiv preprint arXiv:22009.01835}  (\bibinfo{year}{2020}).
\bibitem[{Li et~al.(2012)Li, Luo, Vlasic, Peers, Popović, Pauly, and Rusinkiewicz}]{article}
\bibinfo{author}{H.~Li}, \bibinfo{author}{L.~Luo}, \bibinfo{author}{D.~Vlasic}, \bibinfo{author}{P.~Peers}, \bibinfo{author}{J.~Popović}, \bibinfo{author}{M.~Pauly}, \bibinfo{author}{S.~Rusinkiewicz},
\newblock \bibinfo{title}{Temporally coherent completion of dynamic shapes},
\newblock \bibinfo{journal}{ACM Transactions on Graphics - TOG} \bibinfo{volume}{31} (\bibinfo{year}{2012}) \bibinfo{pages}{1--11}. \DOIprefix\doi{10.1145/2077341.2077343}.
\bibitem[{Wang et~al.(2024)Wang, Wu, Huang, Shi, Shen, Song, Liu, and Li}]{wang2024your}
\bibinfo{author}{F.-Y. Wang}, \bibinfo{author}{X.~Wu}, \bibinfo{author}{Z.~Huang}, \bibinfo{author}{X.~Shi}, \bibinfo{author}{D.~Shen}, \bibinfo{author}{G.~Song}, \bibinfo{author}{Y.~Liu}, \bibinfo{author}{H.~Li},
\newblock \bibinfo{title}{Be-your-outpainter: Mastering video outpainting through input-specific adaptation},
\newblock \bibinfo{journal}{arXiv preprint arXiv:2403.13745}  (\bibinfo{year}{2024}).
\bibitem[{Xu et~al.(2018)Xu, Yang, Fan, Yue, Liang, Yang, and Huang}]{xu2018youtube}
\bibinfo{author}{N.~Xu}, \bibinfo{author}{L.~Yang}, \bibinfo{author}{Y.~Fan}, \bibinfo{author}{D.~Yue}, \bibinfo{author}{Y.~Liang}, \bibinfo{author}{J.~Yang}, \bibinfo{author}{T.~Huang},
\newblock \bibinfo{title}{{YouTube-VOS: A large-scale video object segmentation benchmark}},
\newblock \bibinfo{journal}{arXiv preprint arXiv:1809.03327}  (\bibinfo{year}{2018}).
\bibitem[{Ullah et~al.(2020)Ullah, Jian, Hussain, Guo, Yu, Wang, and Yin}]{ullah2020brief}
\bibinfo{author}{I.~Ullah}, \bibinfo{author}{M.~Jian}, \bibinfo{author}{S.~Hussain}, \bibinfo{author}{J.~Guo}, \bibinfo{author}{H.~Yu}, \bibinfo{author}{X.~Wang}, \bibinfo{author}{Y.~Yin},
\newblock \bibinfo{title}{A brief survey of visual saliency detection},
\newblock \bibinfo{journal}{Multimedia Tools and Applications} \bibinfo{volume}{79} (\bibinfo{year}{2020}) \bibinfo{pages}{34605--34645}.

\end{thebibliography}

\end{document}